# Complexity Issues in Finding Succinct Solutions of PSPACE-Complete Problems

Paolo Liberatore[*]

**Abstract**

We study the problem of deciding whether some PSPACE-complete problems have models of bounded size. Contrary to problems in NP, models of PSPACE-complete problems may be exponentially large. However, such models may take polynomial space in a *succinct representation*. For example, the models of a QBF are explicitely represented by and-or trees (which are always of exponential size) but can be succinctly represented by circuits (which can be polynomial or exponential). We investigate the complexity of deciding the existence of such succinct models when a bound on size is given.

[*]Dipartimento di Informatica e Sistemistica, Università di Roma "La Sapienza", Via Salaria 113, 00198 Roma, Italy `paolo@liberatore.org`

# Contents

# 1 Introduction

Artificial Intelligence (A.I.) is a rich source of hard problems. Many of such problems are complete for the complexity class PSPACE, which contains all problems that can be solved by an algorithm that takes at most a polynomial amount of memory [GJ79, Joh90]. Most of the A.I. problems that are in PSPACE-complete can be recast in one of the two following forms (some exceptions exist, and not all problems in the following forms are in PSPACE-complete).

1. find a tree of polynomial height satisfying a given property;

2. find a sequence of exponential length satisfying a given property.

The first class contains for example the problem of deciding the validity of a Quantified Boolean Formula (QBF). A QBF formula is indeed valid if there exists an and-or tree that expresses the assignments of the variables that make the formula satisfied.

The second class contains for example the problem of deterministic planning in STRIPS. It is indeed well known that there exists instances of the planning problem in STRIPS whose solutions (plans) are all of exponential length. This is what indeed makes planning a PSPACE-hard problem: if the size of the plan is "polynomially bounded", the problem becomes NP-hard.

In both cases the problem can be expressed as: find a "solution" (a suitable data structure) that satisfies a given property. Contrary to the case of NP-complete problems, however, these solutions can be exponentially larger than the problem instance. In particular, some problems of the first kind only admit exponentially large trees as solutions, and some problems of the second kind only admit exponentially long sequences as solutions.

The size of these solution can sometimes be greatly reduced using a succinct representations of trees and sequences. It is indeed easy to show examples of data structures that can be represented very succinctly (with exponentially less space) using an "intelligent" representation. For example, the sequence $0, 1, 2, 3, \ldots, 2^n$ is exponentially long in $n$, but can be easily represented in polynomial space. Indeed, compare the following two representations of the same sequence.

**Explicit representation:**

```
0 1 2 3 4 5 6 7 8 9 10 11 12 13 14 15 16 17 18 19 20 21 22
23 24 25 26 27 28 29 30 31 32 33 34 35 36 37 38 39 40 41 42
43 44 45 46 47 48 49 50 51 52 53 54 55 56 57 58 59 60 61 62
63 64 65 66 67 68 69 70 71 72 73 74 75 76 77 78 79 80 81 82
83 84 85 86 87 88 89 90 91 92 93 94 95 96 97 98 99 100 101
102 103 104 105 106 107 108 109 110 111 112 113 114 115 116
117 118 119 120 121 122 123 124 125 126 127 128
```

**Intelligent representation:**

```
int main() {
  int i;

  for(i=0; i<=128; i++)
    printf("%d ", i);
}
```

Since the program allows determining the sequence uniquely, we can still say that the above 5-line program is a representation of the sequence. It is however much smaller. Other intelligent representations of the same sequence exists. Such representation are called *succinct*, or *compact*. In general, a succinct representation exploits some kind of regularity of the sequence.

Intelligent representations of trees exists as well. For example, the complete tree of 100 levels, with the same labels for all nodes, takes an amount of space greater than $2^{100}$ to be explicitely represented, but can also be represented by a very simple C program.

The question of whether all data structures can be compactly represented has an easy negative solution by a simple counting argument by Shannon [Sha49]: in order to represent all binary strings of length $n$, at least $n$ bits of data are necessary. As a result, some exponential-size solutions of a given problem can be succinctly represented, while others cannot. Moreover, a given problem can have solutions that can be compactly represented while others do not. Therefore, it makes sense to establish whether the solutions of a given problem instance can be succinctly represented or not.

This is especially important because we are often interested in finding solutions even if their explicit representation is exponentially long. Two-player games are an example: a typical solution is how one of the player should move in order to win. Since the moves from the second on depends on the move of the opponent, a solution can be seen as a tree. Yet, such a tree may be representable in little space in some compact form. If such a compression cannot be done, the tree itself cannot be found in full before the start of the game.

Succinct representation of exponentially long sequences may or not be of interest as solutions of problems. In the domain of planning, for example, a plan of exponential length may not be of interest at all because it takes too long to be executed. However, plans of exponential length may be of interest in some cases. As an example, consider the problem of checking whether a system may reach a "forbidden state". Such a problem may be formalized as planning, where plans represent possible evolutions of the system that lead to a forbidden state. Exponentially long plans are then of interest, as they prove that the forbidden state can be actually reached, i.e., the system does not meet the requirement. As a result, the plan is of interest even if it is composed of an exponential number of steps. On the other hand, in order to actually use the given plan (for example, for checking why the system did not meet the specification) it must be represented in an affordable quantity of space. Since we cannot store, or otherwise work with, an exponentially large data structure, then we may only interested into plans that are representable in polynomial space.

Regardless of whether we want to find a sequence or a tree, the problem of checking the existence of solutions that takes a given amount of storage space in succinct form is therefore of interest. We study the complexity of this problem.

We consider some prototypical problems of the two categories: QBF for trees and planning and LTL for sequences. Trees used for QBFs admit two equivalent natural representation, while many exist for the sequences; in particular, the sequences used for representing plans admit at least four succinct representations.

The analysis proceed as in the schema of Figure 1. First, we consider the general problem of compactly representing a tree or a sequence, and compare the space efficiency of the different representations. We then specialize the definitions to the case in which trees or sequences are solutions of a problem. We then show that the problem of finding a solution is different from that of finding a succinct solution of bounded size. We than analyze the complexity of the latter problem. As the complexity may depend on whether the bound is in unary or binary notation, we consider both cases. Namely, we show that finding models of a given size may or may not have the same complexity of the unrestricted problem, and complexity also depends on the specific representation we use for models.

From now one, for the sake of simplicity, we abbreviate "solution in a succinct form" as "**succinct solution**".

---

**Succinct solutions:** Find a succinct representation of solutions;

**Bounding matters:** prove that a bound on the size of the solutions actually makes the problem different; this is done by showing an instance that only has solutions of exponential size (it may be already known, or easy to show from known results);

**Unary $k$:** find the complexity of finding a succinct solution of size $\leq k$, where the integer $k$ is in unary notation;

**Binary $k$:** the same, with $k$ in binary.

---

Figure 1: The steps of the analysis.

Formally, we express the polynomiality of the solution to be found by the notation (unary or binary) we use for the size bound $k$. The unary notation expresses what is often informally stated as "find something that has size polynomial in the size of the instance". Indeed, if $k$ is represented in unary, then any solution of size bounded by $k$ has linear size in the size of the input, since $k$ is part of the input, and it takes $k$ bits to be represented. Therefore, the case of unary $k$ is of interest when the only admissible solutions are those of polynomial size. While this is probably the most interesting case, a grown in size may be allowed, so exponential solutions are also interesting.

Finally, we remark that the problems considered here are taken as prototypical examples of the PSPACE-complete problems that can be expressed as the search for a tree or a sequence, but many other ones exist.

## 2 Preliminaries

We first define the problems that are analyzed in this paper, and then present our notation for boolean circuits, which are the basis for the succinct representations considered in this paper.

## 2.1 QBF, Planning, and LTL

A Quantified Boolean Formula (QBF) is a propositional formula whose boolean variables are either universally or existentially quantified. The following is a general form, where $X_i$ and $Y_i$ are set of variables and $E$ is a propositional formula built on these variables.

$$\forall X_1 \exists Y_1, \ldots, \forall X_n \exists Y_n.E$$

The propositional formula $E$ is called the *matrix* of the QBF. We do not impose any special form on the QBF, so the first quantifier can also be existential and the last one universal.

The problem of checking the validity of a QBF is important both theoretically and practically. From a theoretical point of view, the problem of validity of QBFs is a prototypical PSPACE-complete problem; moreover, it is the simplest possible formalization of *games*: a QBF can express whether a player can win. Indeed, this is the case if, for any opponent's move there is a countermove such that, for any opponent's move, etc. Besides games, similar problems of the same structure are MDPs [Bel57], probabilistic and nondeterministic planning [LGM98]. From a practical point of view, QBFs have been used as a common "solution point" for PSPACE problems: as any PSPACE problem can be reduced to validity of QBFs in polynomial time, then a solver for QBF can actually solve, via a translation, any problem that is in PSPACE [CSGG02, Rin99, DLMS02].

The second problem we consider is that of planning. An instance of the problem of planning is composed of an initial state, a specification of the goal states, and a set of actions (often called "operators"). The solution of a planning problem is a plan, i.e., a sequence of actions that leads from the initial state to a goal state. The planning formalism can also express some problems of verification: in order to check that a system never goes into a "forbidden state", we can instead try to plan to reach such states. If a plan exists, the system does not met the requirement.

Instead of using common planning formalisms such as STRIPS [FN71] or ADL [Ped89], or even enhanced action description languages such as $\mathcal{A}$ [GL93], we simply assume that actions are described by polynomial functions that determine the next state from the state in which they are applied (non-applicability of actions is not relevant.) This way, we are at the same time generalizing the formalisms for (deterministic and non-probabilistic) planning, without losing the property of actions being simple to evaluate.

LTL is a temporal logic whose underlying model of time is linear. A model of LTL is a Kripke stricture where the possible worlds are connected in a chain. The model operators are four: the three unary operators **X**, **F** and **G** and the binary model operator **U**. Their intended meanings are:

- **X** is "next": $\mathbf{X}\alpha$ means that $\alpha$ is true in the next state;

- **F** is "eventually": $\mathbf{F}\alpha$ means that $\alpha$ is true in some future state;

- **G** is "always": $\mathbf{G}\alpha$ means that $\alpha$ is true in all future states;

- **U** is "until": $\alpha\mathbf{U}\beta$ means that $\alpha$ is true until $\beta$ becomes true.

Therefore, $\mathbf{X}x$ is true in a point of the chain if $x$ is true in the *next* state, $\mathbf{F}x$ is true in a state if $x$ is true in some successive state of the chain, and $\mathbf{G}x$ is true if $x$ is always true from this point on.

The following figure shows the semantics of $a\mathbf{U}b$: the $b$ is true in some successive state and $a$ is true until then. The same holds for $A\mathbf{U}B$ where $A$ and $B$ are arbitrary formulae.

We omit the formal definition of the semantics of LTL [Gol87].

## 2.2 Succinct Representations

Every data structure can be represented in more than one way. In this paper we are interested into two simple data structures, sequences and trees. By "explicit representation" of a data structure we mean the data structure itself. As a result, the explicit representation of a sequence is the sequence and the explicit representation of a tree is the tree.

The lenght of a sequence is the number of elements it contains. However, some sequences allows for a shorter representation of them. For example, the sequence composed of one thousand identical elements have lenght 1000, but one can think of several other ways for representing such a sequence in a very short space. The same holds for trees: the complete tree of one thusand levels has $2^{1000} - 1$ nodes, but can be represented in a very compact form.

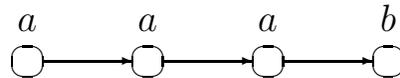

Figure 2: A model in which $a\mathbf{U}b$ is true: the variable $b$ will eventually be true and the variable $a$ is true in all points before this happens.

In this section we make some general considerations about succinct represention. We can formally define a representation of a class of data structures as follows.

**Definition 1** *A representation of a class of data structures $D$ is a surjective function $r : D' \to D$ whose domain $D'$ is another class of data structures.*

The idea is that every element of $D$ is represented by at least one element of $D'$. This is what "surjective" means: for every $d \in D$, there must exists a least one $d' \in D'$ such that $r(d') = d$. This is exactly what we require for a representation of the elements of $D$: every element of $D$ must be represented by an element of $D'$. Still better, it must be represented by *at least* an element of $D'$. The function $r$ allows determining the represented element $d = r(d')$ from its representation $d'$.

The definition of representations already contains the first requirement on representations: every element of the set of data structures $D$ we consider must be "representable", that is, there is an element of $D'$ that represents it. The second requirement we consider is about the time needed to determine the represented element from its representation.

**Definition 2** *A representation $r$ of a class of data structures $D$ is polynomial if $r$ is input-output polynomial-time.*

In other words, for every $d' \in D'$, it must be possible to reconstruct the represented data structure $d = f(d')$ in time that is polynomial in the size of $d$ plus the size of $d'$. Using both the size of $d$ and of $d'$ is necessary because large elements of $D$ can be represented by small elements of $D'$, and a large element of $D$ can represent a small element of $D$. Both cases happen in the compact representation we use.

An example of a representation of a sequence is a device that allows determining the next element given the previous ones. The requirement of polynomiality imposes that the device can only take a polynomial amount of time to produce a result. The most general devices that run in time polynomial in the combined size of their inputs and outputs are the boolean circuits.

Indeed, every polynomial-time function can be expressed as a family of circuits of polynomial size. More precisely, every function $f : \{0,1\}^n \to \{0,1\}^m$ can be expressed by a circuit having $n$ input gates and $m$ output gates, and vice versa. If the function $f$ can be evaluted in polynomial time, the corresponding circuit has polynomial size, and if the circuit has polynomial size the corresponding function is polynomial-time. Therefore, polynomial-time functions correspond to polynomial-size circuits and vice versa.

Circuits can be defined in several equivalent ways, for example using graphs. We however prefer the definition based on propositional formulae because is it easier to write in a text, and more coherent with the rest of the formal part of the paper. A circuit with $n$ inputs and $m$ outputs is a boolean formula containing three sets of variables:

**input variables:** $X = \{x_1, \ldots, x_n\}$;

**output variables:** $O = \{o_1, \ldots, o_m\}$;

**internal variables:** $Z = \{z_1, \ldots, z_k\}$.

A circuit $C$ is a formula made of a conjunction of parts as follows, where each of $C_i$ and $G_i$ is a boolean formula of some of its arguments containing a single boolean operator (and, or, not).

$$C = \bigwedge_{i=1}^{k} \left( \begin{array}{rcl} z_1 & \equiv & C_1(x_1, \ldots, x_n) \\ z_2 & \equiv & C_2(x_1, \ldots, x_n, z_1) \\ z_3 & \equiv & C_3(x_1, \ldots, x_n, z_1, z_2) \\ & \vdots & \\ z_k & \equiv & C_k(x_1, \ldots, x_n, z_1, \ldots, z_{k-1}) \end{array} \right) \wedge \bigwedge_{i=1}^{m} o_i \equiv G_i(x_1, \ldots, x_n, z_1, \ldots, z_k)$$

In words, the value of each $z_i$ is determined in a simple way from the value of the input variables and of the preceeding $z_j$ with $j < i$. The output variables values are determined from the values of the variables $x_i$ and $z_i$. Other definitions of circuits exist, but are all equivalent to the one given here. The point is simply that there is a set of input variables, whose value allow for determining the value of other variables, which in turns allows for determining the value of the output variables.

The value of the output variables can be determined from the value of the input variables in linear time (linear in the size of the circuit). A converse result is also easy to give: any polynomial-time function $f : \{0,1\}^n \to \{0,1\}^m$ can be expressed as a polynomially sized circuit with $n$ input and $m$ outputs. Therefore, circuits are a good way for representing polynomial-time functions.

When studying QBFs, it will be important to embed a circuit into a QBF. In particular, it will be useful to use a circuit to encode a relationship between input and output variables that are both quantified in the QBF. Some problems can indeed be expressed in a form like:

$$\forall X_1 \exists Y_1 \ldots \exists Y_n (\text{a property is true})$$

and the property can be checked in polynomial time from the value of the variables $X = X_1 \cup Y_1 \cup \cdots \cup Y_n$. Since we can verify this property in polynomial time, we can express it as a circuit with input variables $X = X_1 \cup Y_1 \cup \cdots \cup Y_n$. Since a circuit is a formula of a particular kind, we can turn the above into a real QBF.

The property can only be true or false. As a result, its corresponding circuit has a single output variable $o_1$. We denote by $Z$ the internal variables of the circuit. The property is true if and only if $o_1 \wedge F$ is true. In order to express this formula into a closed QBF, we must specify whether $o_1$ and $Z$ are existentially or universally quantified. Only the existential quantification leads to the correct result. Indeed, the output value of the circuit is what results from checking the existence of a value for the output and internal value that satisfy the circuit. As a result, the above "informal formula" is equivalent to the following QBF:

$$\forall X_1 \exists Y_1 \ldots \exists Y_n \exists o_1 \exists Z \ . \ o_1 \wedge C$$

Formally, we can state the following theorem.

**Theorem 1** *A property expressed as:*

> *for all values of $X_1$ there exists a value of $Y_1$ such that ... the property $P$ is true*

*such that the property $P$ can be checked in polynomial time from the value of $X_1 \cup Y_1 \cup \ldots$, is valid if and only if the following QBF is valid:*

$$\forall X_1 \exists Y_2 \ldots \exists o_1 \exists Z \ . \ o_1 \wedge C$$

*where the formula $C$ is the circuit expressing the property $P$, the variable $o_1$ is its only output, and $Z$ is the set of its internal variables.*

The existential quantification of the variables $o_1$ and $Z$ can be turned into a universal quantification by replacing the circuit $C$ with the following one:

$$U(C) \;=\; \bigvee_{i=1}^{k} \left( \begin{array}{rcl} z_1 & \not\equiv & C_1(x_1,\ldots,x_n) \\ z_2 & \not\equiv & C_2(x_1,\ldots,x_n,z_1) \\ z_3 & \not\equiv & C_3(x_1,\ldots,x_n,z_1,z_2) \\ & \vdots & \\ z_k & \not\equiv & C_k(x_1,\ldots,x_n,z_1,\ldots,z_{k-1}) \end{array} \right) \vee (o_1 \not\equiv G_1(x_1,\ldots,x_n,z_1,\ldots,z_k))$$

The output value is now the (only) value of $o_1$ that satisfy the formula $\forall Z.U(C)$. Indeed, we have to consider all values for each variable of $z_i$. However, when a variable $z_i$ does not have the same value as the corresponding $C_i(\ldots)$, the formula $U(C)$ is true. The only remaining case that is important for the satisfiability of the formula is when each $z_i$ has the same value of $C_i(\ldots)$. In particular, the output is true if and only if the QBF $\forall o_1 \forall Z \,.\, \neg o_1 \vee U(C)$ is valid: this formula is indeed valid only if $U(C)$ is consistent with $o_1$ being true.

Intuitively, a circuit is a quantified boolean formula; however, being of a special form that is polynomial to evaluate, the kind of quantifiers we use is not important.

Summarizing, an informal property like $\forall X_1 \ldots \exists Y_n.$(property) involving quantifications can be turned into a formal QBF by replacing the property with a QBF that contains only universally quantified variables. Formally, this can be stated as in the following theorem.

**Theorem 2** *A property expressed as:*

> *for all values of $X_1$ there exists a value of $Y_1$ such that ... the property $P$ is true*

*such that the property $P$ can be checked in polynomial time, is valid if and only if the following QBF is valid:*

$$\forall X_1 \exists Y_1 \ldots \forall o_1 \forall Z \,.\, \neg o_1 \vee U(C)$$

*where the formula $C$ is the circuit expressing the property $P$, the variable $o_1$ is its only output, and $Z$ is the set of its internal variables.*

# 3 Succinct Trees

Trees can be compactly represented in several ways. The general requirements of a representation are:

1. every tree can be representable;

2. an explicit representation of the tree can be reconstructed from the succinct representation in time polynomial in the size of the explicit representation.

Which representation is the best one depends on how trees are used. Namely, a tree may be visited from the root to the leaves, from a leaf to the root, or level-by-level. Depending on how the tree has to be used, a representation may be more or less natural. We consider the two following representations.

**Directional Representation:** a circuit that, given a path from the root to a node, tells the children of the node;

**Positional Representation:** this is based on the representation of trees using array; namely, it consists in a circuit that takes as input an integer $i$ and tells the node in position $i$ of the array; in other words, this is a succinct representation of an array.

The trees that are solutions of the problems of QBF and nondeterministic planning appear to be better represented in the first way. Indeed, a QBF model should tell the value of the existential variable given the preceeding ones, and the directional representation tells exactly what is the step to do after some variables have been assigned. The same is true for nondeterministic planning, where the solution tells the action to take given the previous ones.

The second representation is more useful when trees have to be visited in some other way, for example by levels. Indeed, this representation allows for finding the nodes at some level quite easily. It can be easily shown that these two representations are indeed equivalent.

**Theorem 3** *Any directional representation of a tree can be translated into a positional representation of the same tree of polynomial size, and vice versa.*

*Proof.* Given a directional representation of a tree, we can find the node in some position by simply starting from the root and generating the path leading to the node in the given position.

On the other way around, given a sequence of nodes, we can find ithe children of the last one, given the positional representation of the tree by simply determining the position of the last node of the path. □

Sequences and trees can be put in correspondence in several ways. For example, a sequence can be represented by the nodes of the tree (in the order given by the vector representation) of by the leaves. In order to represent tree with sequences, however, only the first method can be used, and the fact that some positions of the tree may be empty has to be taken into account. We not analyze this correspondence any further.

## 3.1 QBF Model Finding

### 3.1.1 Succinct Models

The problem we consider is the validity of a QBF: deciding whether a given formula $\exists X_1 \forall Y_1, \ldots, \exists X_n \forall Y_n.E$ is valid. A model of such a QBF is an and-or tree in which *and* nodes correspond to universal variables, while *or* nodes correspond to existential variables. Such a tree allows determining the value of the existentially quantified variables given the value of the variables in the preceeding quantifications.

Since all internal "and" nodes have two children, the size of a QBF model is exponential in the number of universally quantified variables. Models can however sometimes be more compactly represented using the representations of tree defined in the previous section: directional or positional.

The directional representation appears to be the most direct in representing models of a QBF. Indeed, a model of a QBF is a way for determining the value of the existential variables from the value of the previous variables. In terms of trees, each node is determined from the path linking the root to it. The directional representation of a tree is exactly a way for representing trees using a function for telling a node from the path from the root to it.

In the directional representation of trees, a QBF model is a circuit rep-reseting a function from paths to nodes. Formally, a QBF model is a set of functions determining the values of each set of existentially quantified variables from the preceeding universal ones. The compact representation of a model of a QBF $F = \forall X_1 \exists Y_1 \ldots \forall X_n \exists Y_n \, . \, E$ is a set of $n$ circuits $M = \{C_1, \ldots, C_n\}$. Each circuit $C_i$ has $X_1, \ldots, X_i$ as inputs and $Y_i$ as output.

Evaluating a formula $F = \forall X_1 \exists Y_1 \ldots \forall X_n \exists Y_n \,.\, E$ w.r.t. such a model $M = \{C_1, \ldots, C_n\}$ is done as follows: consider each possible evaluation of the variables in $X_1$, and determine the corresponding value of $Y_1$ with the circuit $C_1$; then, consider any possible evaluation of $X_2$, and for each of them determine the value of $Y_2$ from those of $X_1$ and $X_2$ using the circuit $C_2$; when all variables are assigned, check whether the truth evalution of the variables satisfies the formula $E$. This recursive procedure determines in polynomial space whether a model satisfies a formula.

A trivial consequence of this algorithm is that the problem of checking whether a QBF has a model that is large at most $k$ is in PSPACE, if $k$ is represented in unary notation: nondeterministically guess a set of circuits $M$ of size bounded by $k$, and then check the formula. Since checking only requires polynomial space, the whole problem is $\text{NP}^{\text{PSPACE}}$, which is known to be equal to PSPACE [Sav70].

A compact representation of a model, if any, can also be determined by a QBF prover that returns an evaluation of the first existentially quantified variables. Indeed, we can express the existence of a compact model of a QBF as a guessing followed by a PSPACE problem, and this means that what is guessed (the compactly represented model) is encoded in the first existentially quantified variables.

The other problems that are closely related to QBFs, such as MDPs, would also lead to a similar representation of models. The corresponding concept of models in MDPs is that of policy: a policy is a way for telling the action to execute in each possible state, that is, determining what we can choose (the action, or the existential variables) from what we cannot choose (the state we are currently in, or the value of universal variables.)

### 3.1.2 Bounding Matters

Finding a succinct representation of models is the first step of the analysis as shown Figure 1. For the QBF validity problem, we have chosen the directional representation in the previous section. The second step is that of proving that bounding matters: we have to show that the problem of finding a model is affected by the presence of a bound in the size of the model.

We first show that bounding the size of models modify the problem of QBF validity. In other words, we prove that the validity of a QBF can change if we add a constraint on the size of its models. The formula $\forall x_1 \exists y_1 . x_1 \equiv y_1$ has exactly one model $\{y_1 \equiv x_1\}$: this is the model in which $y_1$ is assigned to the same value of $x_1$. Nevertheless, it has no model of size bounded by $k=0$, as the only two such models are $\{\top\}$ and $\{\bot\}$, and none of them satisfy the QBF.

This very argument can be used to prove that constraining models to be representable with a given size makes valid formulae unsatisfiable.

**Theorem 4** *For any $k$ there exists a valid QBF that have size polynomial in the value of $k$, and have no models representable in size $k$ or less.*

*Proof.* The QBF of the theorem is the following one:

$$F = \forall x_1 \exists y_1 \cdots \forall x_k \exists y_k \, . \, y_k \equiv x_1 \vee x_2 \vee \cdots \vee x_k$$

In any model $\{C_1, \ldots, C_k\}$ of this formula, the last circuit $C_k$ must be equivalent to $y_k \equiv x_1 \vee \cdots \vee x_k$. This circuit alone has size $k+1$, and no equivalent circuit is smaller. This proves that $F$ is valid, but has no model of size $k$ or less. $\square$

This theorem proves that the problem of establishing the validity of a QBF and the problem of establishing the validity of a QBF with a model of a given size are different problems. Nevertheless, the claim itself was quite obvious from the beginning. What would be more interesting to prove is that there are valid formulae that do not have any polynomially sized models.

In the other way around: is the sentence "a formula is valid if and only if it has a polynomial model" true for all formulae? Such property would be similar to the Finite Model Property. We formally express the question for arbitrary problems as follows.

**Property 1 (Polynomial Solution Property)** *A problem has the Polynomial Solution Property if there exists a polynomial $p$ such that any problem instance $I$ has a solution if and only if it has a solution of size bounded by $p(||I||)$.*

This property can hold or not depending on what we consider the solution of a problem to be. For example, the QBF validity problem has the polynomial solution property if we take the solution of the problem to be only the validity of the problem, since this solution can only be "yes" or "not".

If the solution of the QBF validity problem is instead a model of the QBF, the problem will be proved not to have the polynomial solution property. In general, the property might hold or not depending on how solutions are represented, but this is not the case for the QBF validity problem.

We show that the Polynomial Solution Property does not hold for the QBF validity problem when the solution is a model of the QBF in the directional representation, unless the polynomial hierarchy collapses. The proof is not much different from the ones used for POMDPs by Papadimitriou and Tsitsiklis [PT87].

**Theorem 5** *If the Polynomial Solution Property holds for the QBF validity problem when the solution is a model in the directional representation, then $\Pi_2^p \subseteq \Sigma_2^p$.*

*Proof.* Let us consider the $\Pi_2^p$-hard problem of checking the validity of QBFs in the form $F = \forall X \exists Y . E$. We prove that the Polynomial Solution Property implies that this problem can be expressed as a QBF formula in which the existential quantifiers are before the universal one. Since the validity of such formulae is in $\Sigma_2^p$, this proves the claim.

Assuming the Polynomial Solution Property, any QBF is valid if and only if it has a model of polynomial size. Let $p$ be this polynomial: $F$ is valid if and only if there exists a model $M$, of size bounded by $p(||F||)$, that expresses the value of the existentially quantified variables in function of the universal ones.

Therefore, we can express the validity of $F$ as follows:

> There exists a model $M$ such that, for each possible value of $X$, the value of $Y$ that result from evaluating $M$, together with the value of $X$, satisfy $E$.

The property "the value of $Y$ ... satisfy $E$" can be checked in polynomial time, as it amounts to determining the output of the circuits in $M$, and then using this output to check the validity of a propositional formula $E$. Note that the polynomiality of $M$ is essential both to express the model as a set of propositional variable and to make the property above polynomial.

By Theorem 1, the above property can be formalized by the following QBF, where $C$ is the circuit expressing the polynomial property, and $o_1$ is its only output while $Z$ is the set of its internal variables.

$$\forall M \exists X \exists o_1 \exists Z \ . \ o_1 \wedge C$$

This proves that a $\Pi_2^p$-complete problem, as the validity test of a formula like $F$ is, can be polynomially reduced to a problem in $\Sigma_2^p$, and therefore $\Pi_2^p \subseteq \Sigma_2^p$. $\square$

A similar proof can be used to show that the Polynomial Solution Property for the QBF validity problem implies the collapse of PSPACE to $\Sigma_2^p$ as well.

A similar result can be proven for every problem that is PSPACE-hard and for every succinct representation of solution such that checking the validity of a solution over the problem instance is in a class of the polynomial hierarchy. The proof is similar to the one shown above: if the polynomial solution property holds, then the problem can be expressed as: "there exists a valid solution". This sentence can be formally expresses as: "there exists a solution of size bounded by a polynomial in the problem instance, such that the solution is valid w.r.t. the problem instance". If the problem of verifying a solution is in a class $\Pi_n^p$, the problem itself is in the class $\Sigma_{n+1}^p$. Since the problem is PSPACE-hard, that would imply that PSPACE $\subseteq \Sigma_{n+1}^p$.

This result establishes the difference between the problems of existence of model with or without size limits. The problem with no bound has already been investigated, and many results are in the literature [Pap94]. In the next section, we concentrate on the problem with bounds.

### 3.1.3 Complexity: Unary Notation

In this section, we begin the study of complexity of the problem of deciding the existence of a model of size bounded by $k$, corresponding to the third and fourth steps of Figure 1. The first case is when $k$ is in unary. As already remarked, if we can afford saving a model of size $k$, we also have the room to write $k$ in unary. This assumption corresponds to setting a polynomial bound on the size of the models.

**Theorem 6** *The problem of deciding whether a QBF formula has a model of size bounded by $k$ is in $\Sigma_2^p$, if $k$ is represented in unary notation.*

*Proof.* Deciding whether a formula $F = \forall X_1 \exists Y_1 \ldots . E$ has a model of size bounded by $k$ can be informally expressed as follows:

there exists a model $M = \{C_1, \ldots, C_n\}$ such that, for all values of $X_1, X_2, \ldots, X_n$, these values and the values of $Y_i$ that result from evaluating the circuits $C_1, \ldots, C_n$, satisfy the formula $E$.

Note that the fact that $M$ is of size (at most) $k$, makes the first "informal quantification" of the formula to be expressed as the existential quantification of a polynomial number of variables. Moreover, the sentence "these values and ... satisfy the formula $E$" can be then checked in polynomial time.

Using Theorem 2, the above property can be expressed as a QBF. Namely, we can express the property as a circuit $C$, with output $o_1$ and internal variables $Z$. By Theorem 2, the property above can be formalized as:

$$\exists M \forall x_1 \ldots x_n \forall o_1 \forall Z \; . \; \neg o_1 \vee U(C)$$

This QBF has only one alternation of quantifiers; namely, there is a set of existentially quantified variables followed by a set of universally quantified variables. Therefore, the problem is in $\Sigma_2^p$. □

Note that the unary representation of $k$ is essential, as it allows expressing the existence of a circuit with the existence of values for a set of variables $M$ with polynomial cardinality.

**Theorem 7** *Checking existence of a model of size bounded by $k$ is $\Sigma_2^p$-hard, if $k$ is represented in unary notation.*

*Proof.* The validity of a formula of the form $F = \exists X \forall Y \; . \; G$ amounts to checking the existence of a model for it. Such a model is always of polynomial size: since no universal variable precedes the existential ones, the circuit giving the value of the variables $X$ has no input variables, i.e., it is a set of formulae like $x_i \equiv \mathsf{true}$ or $x_i \equiv \mathsf{false}$. Therefore, the problem of checking the existence of a model for a QBF $F = \exists X \forall Y \; . \; G$ is exactly the same as checking the existence of a model of size bounded by $||X||$. As a result, the problem of checking whether a formula has a model of given size is $\Sigma_2^p$-hard. □

### 3.1.4 Complexity: Binary Notation

We now study the complexity of the problem when $k$ is not expressed in unary but in binary notation. The first, obvious result is that the complexity of finding models is at least the same as in the case of unbounded models.

**Theorem 8** *The problem of deciding whether a QBF formula has a model of size bounded by $k$ is* PSPACE*-hard, if $k$ is represented in binary notation.*

*Proof.* Let $n$ be the number of all variables in the formula. Any model is composed of circuits with (at most) $n$ input variables and $n$ output variables. Now, any function from $n$-tuples of bits to $n$-tuples of bits can be expressed as a circuit with an exponential number of gates and no internal variables. Therefore, using $k = c2^n$ with a suitable value of the constant $c$, the bound is made irrelevant, that is, a QBF is valid if and only if it has models of size bounded by $k$. □

The problem could be in PSPACEor be EXPTIME-hard. The latter however implies a separation between P and PSPACE.

**Lemma 1** *If the problem of deciding the existence of a model of a QBF, of size bounded by $k$ (represented in binary) is not in* P*, then* P$\neq$PSPACE.

*Proof.* We prove the claim by inverting it: we assume that P=PSPACE and prove that the problem of existence of models is in P.

We first show that a satisfying evaluation (if any) of the existentially quantified variables $Y_i$ can be determined in polynomial space from the values of the preceding variables. This is true simply because solving the QBF that results from instantiating the values of the preceding variables is in PSPACE.

Since P = PSPACE, this problem is in P. As a result, there exists a circuit of polynomial size that tells a satisfying evaluation (if any) of $Y_i$ having as input the values of the preceding variables. Since a model of the QBF in succinct representation is the set of these circuits, for every valid QBF there exists a model that can be succinctly represented in polynomial space.

As a result, we can solve the problem of the validity of a QBF by guessing a model and then checking its validity. Again, the validity of a model can be checked in polynomial space, and is therefore in P. This algorithm shows that the problem of validity of a QBF is in NP, and is therefore in P as well, since P=PSPACE. The same procedure can be used for the case in which there is a bound $k$ on the size of the model. Indeed, every QBF is valid if and only if has a model of polynomial size. If $k$ is greater than this polynomial, the bound does not affect validity. If $k$ is less than this polynomial, then the algorithm of guess-and-check above can still be used because the model to be guessed has size polynomial. □

Since EXPTIME-hard problems are known not to be in P, the problem of checking the existence of a succinct model of size $k$ in binary is not EXPTIME-hard unless P$\neq$PSPACE.

**Corollary 1** *If the problem of deciding the existence of a model of a QBF of size bounded by $k$ (represented in binary) is* EXPTIME-*hard, then* P$\neq$PSPACE.

Whether the problem is in EXPTIME is still an open problem.

## 4 Succinct Sequences

The problem we consider is that of representing sequence of elements. For the problem of planning, a solution can be a sequence of states or a sequence of actions. A sequence of actions generates a single sequence of states, but the same sequence of states can be generated by many sequences of actions. The way models are represented depends on two factors:

1. the sequences can contain repeated elements or not;

2. the set of all possible elements of the sequences is given explicitly or it is the set of propositional interpretations over a set of boolean variables.

For the problem of planning, a solution can be a sequence of states or a sequence of actions. These two sequences are exactly the opposite of each other as for the two properties above.

A sequence of states cannot (or can be supposed not to) contain repetitions: once a goal state is reached, there is no need to go ahead. The states are propositional interpretations over a set of propositional variables.

A sequence of actions can contain repeated elements: in some cases, the only plans of an instance contain the same action more than once, or even an exponential number of times. The set of actions is typically given explicitly, that is, an enumeration of all possible actions is given. Sequences of actions are therefore exactly the opposite of sequences of states w.r.t. the properties of repeated elements and representation of the elements.

A succinct representation that can be used for every sequence is that of a circuit producing the $n$-th element of the sequence having $n$ has input. We call this the time/element representation of the sequence.

**Definition 3 (Time/Element Representation)** *The time/element representation of a sequence is a pair $\langle C_{TE}, N \rangle$, where $N$ is the length of the sequence and $C_{TE}$ is a circuit that outputs the n-th element of the sequence whenever its input is n.*

A sequence with no repetitions can also be represented by a rule telling the next element from the previous one.

**Definition 4 (Next Element Representation)** *The next element representation of a sequence with no repetitions is a triple $\langle s_0, C_{EE}, N \rangle$, where $s_0$ is the first element of the sequence, $N$ is the length of the sequence and $C_{EE}$ is a circuit that outputs the $n+1$-th element of the sequence whenever its input is n-th element.*

The two representations to solutions can be applied to the the planning problem as follows. The solution of a planning problem is either a sequence of states or a sequence of actions. Both kinds of sequences can be represented both in the time/element representation and in the next element representation, the latter requiring some care for the case of sequences of actions.

**Time/State Representation:** a sequence of states in the time/element representation;

**Next State Representation:** a sequence of states in the next element representation;

**Time/Action Representation:** a sequence of actions in the time/element representation;

**Next Action Representation:** since a sequence of actions can contain repeated elements, the element/element representation cannot be used directly; however, since the sequence of states cannot contain repeated element, a sequence of actions can be represented by a correspondence between every state and the action to execute in it.

In the next section, we show some general results about the succinctness of these four representations of sequences.

## 4.1 Succinct Sequences of States

In order to prove that some representations of sequences are more succinct than other ones, we employ concepts from computational complexity. In particular, we rely on the existence of polyomial-time one-way functions, and use the concepts of compilability classes and reductions [CDLS02, Lib01]. The compilability classes can be applied to the problem of succinct representations as follows. If $X$ and $Y$ are two methods for representing the same data, the following strategy proves that $Y$ can represent some sequences in exponentially less space.

1. Choose a problem whose input is a sequence, and possibly some other data;

2. prove that this problem can be easily solved when the sequence is in representation $X$;

3. prove that the problem is hard when we use representation $Y$.

When "compilability hardness" is used (instead of the hardness based on polynomial many-one reductions), we can conclude that there are sequences that can be represented in polynomial space using representation $Y$, but always takes super-polynomial space in representation $X$. Note that the more "space efficient" representation is the one that is the more complex. This tradeoff between complexity and space efficency seems to be a general phenomenon [CDS96].

We first consider the two representations of sequences of states: the time/state and the next state representations, which are formally defined as follows.

**time/state representation:** a sequence of states is a pair $\langle C_{TS}, N \rangle$ where $N$ is the number of elements of the sequence and $C_{TS}$ is a circuit that outputs the $t$-th element of sequence when its input is $t$;

**next state representation:** a sequence of states is a triple $\langle s_0, C_{SS}, N \rangle$ where $N$ is the number of elements of the sequence, $s_0$ is the first element, and $C_{SS}$ is a circuit that outputs the next state when its input is the current one.

We now formally prove the following result: there are sequences that take polynomial space in the time/state representation, but cannot be polynomially represented in the next state representation. The first step of the strategy is to choose a problem.

**Definition 5** *The time point problem is that of deciding whether a state is the t-th element of a sequence.*

Since this problem instances contain a sequence (plus the number $t$) it satisfies the only requirement of the strategy.

The second step is to prove that the problem is easy to solve if the sequence is in one representation. The time point problem is trivially polynomial if the sequences are in the time/state representation, as it amounts to determining the output of the circuit $C_{TS}$ given its input.

The last step is to prove that the problem is hard if the sequences are in the other representation. We prove that the time point problem is $\|\!\sim$NP-hard, when the next state representation is used.

**Theorem 9** *The time point problem is $\|\!\sim$NP-hard, if the fixed part is the next state representation of the sequence and the varying part is the time point.*

*Proof.* We give a reduction from ∗3sat to the time point problem. We assume that formulae in 3cnf are uniquely associated to strings of $m$ bits. We also consider propositional interpretations as strings of $r$ bits.

The reduction is as follows. Given a formula, the states of the sequence are composed of four parts, like $\langle F, M, x, y \rangle$, in which $F$ is a string of $m$ bits (and is thus associated to a formula), $M$ is a string of $r$ bits (and is thus associated to a propositional interpretation), and $x$ and $y$ are bits.

The next state circuit we consider is $C_{SS}(\langle F, M, x, y \rangle) = \langle F', M', x', y' \rangle$, where:

- $F' = F$ if $y$ is 0; otherwise, $F'$ is the string of $m$ bits that follow $F$ (i.e., $F' = F + 1$);

- $M' = M + 1$ if $y = 0$; otherwise $M'$ is the string of all 0's;

- if $M$ satisfies $F$ then $x'$ is 1; otherwise $x' = x$;

- if $M$ is composed only of 1's, then $y' = 1$; otherwise, $y' = 0$.

The represented sequence is composed of a separate chunk for each formula. The first state of the chunk is $\langle F, 0\ldots 0, 0, 0\rangle$. The next state contains $F$, the next model, and $x$ is determined according to the validity of $F$ in the first model. The sequence ends when all models have been considered. Indeed, when the string representing the model is $1\ldots 1$, in the next state $y$ is set to 1; in the next state, a new formula is considered, and the model is reset to the one setting to 0 all variables. The following table gives an idea of the chunk corresponding to a formula $F$ whose first satisfying model is $01101\ldots 01$.

| Formula ($m$ bits) | Model ($r$ bits) | $x$ | $y$ | |
|---|---|---|---|---|
| $F'$ | $00000\ldots 00$ | 0 or 1 | 1 | |
| $F$ | $00000\ldots 00$ | 0 | 0 | |
| | $\vdots$ | | | |
| $F$ | $01101\ldots 01$ | 0 | 0 | first satisfying model of $F$ |
| $F$ | $01101\ldots 10$ | 1 | 0 | in the next state, $x$ goes to 1 |
| | $\vdots$ | | | |
| $F$ | $11111\ldots 11$ | 1 | 0 | |
| $F$ | $00000\ldots 00$ | 1 | 1 | |
| $F''$ | $00000\ldots 00$ | 0 | 0 | |

The first element of the table is the last state of the chunk of the previous formula $F'$. The first element of the chunk associated to the formula $F$ is the state composed of $F$, the sequence of $r$ zeros, and two other 0's. The successor of each state has the next string of $r$ bits. The bit $x$ is set to 1 whenever the model associated to the string of $n$ bits satisfies the formula. The last string of $r$ bits it that composed of all 1's. The successor of this state is the one in which $y$ is finally set to 1. This concludes the chunk: the next element of the sequence is the first state of the chunk associated to the formula $F''$, which is the one that follows $F$ in our representation.

It is now clear how satisfiability of $F$ can be checked. The first element of the sequence is the first state of the chunk associated to the first formula. It is then easy to locate the chunk corresponding to a formula in the whole sequence. Indeed, each chunk is composed of $2^r + 1$ states. Therefore, the chunk that corresponds to the formula whose string of bits represent the number $k$ starts at the $k \times (2^r + 1) + 1$-th element of the sequence. The *last* element of the chunk indicates whether the formula is satisfiable or not (the bit $x$ tells this). Therefore, given a formula which is represented by the string of $m$ bits whose represented integer is $k$, its satisfiability amount to checking whether the $(k+1) \times (2^r + 1)$-th element of the sequence is $\langle k, 0 \ldots 0, 1, 1 \rangle$.

This proves that the problem of satisfiability can be polynomially reduced to the problem of determining the $t$-th element of a sequence in the next state representation. Moreover, the circuit and the initial state only depends on the number of variables in the formula, while the formula only affects the specific integer $t$ and state. This is therefore a $\leq_{nucomp}$ reduction from $*$3sat to the time point problem, which is thus $\|\leadsto$NP-hard. □

This proves that the next state representation is more space efficient than the time/state representation on some sequences.

**Corollary 2** *Some sequences can be represented in polynomial space in the next state representation, but cannot in the time/state representation.*

We indicate such a result by **SS↛TS**, or graphically as follows.

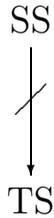

The fact that some sequences are better represented in the next state representation does not make it the best representation for all sequences. Indeed, there may be other sequences that are shorter in the time/state representation. There is some evidence that this is actually the case: we show that the possibility of translating a time/state representation into a next state one in polynomial time implies the invertibility of all polynomial permutations.

**Theorem 10** *If translating from a time/state representation of a sequence into a next state one is polynomial-time, then any polynomial permutation is invertible in polynomial time.*

*Proof.* A permutation is a one-to-one length-preserving function (i.e., a mapping from strings of $n$ bits into strings of $n$ bits that always maps different strings into different strings). Let $f$ be a polynomial permutation.

We consider the sequence represented (in the time/state representation) by $\langle C_{TS}, 2^{n+1}\rangle$, where:

$$C_{TS}(x0) = f(x)0$$
$$C_{TS}(x1) = x1$$

This circuit is polynomially large, as $f$ is a polynomial function. If this circuit can be translated into a next state circuit $C_{SS}$ in polynomial time, then the permutation $f$ is invertible in polynomial time: given $x$, compute the circuit $C_{TS}$ for the given size of $x$, translate it into $C_{SS}$, and then compute $x1 = C_{SS}(f(x)0)$. □

The invertibility of all polynomial-time permutations seems unlikely [RH02, HT03]. Note, however, that this theorem is about the possiblity of translating from the time/state into the next state representation in polynomial time. In other words, it is about the time complexity of the translation, not about the size of its output. The problem of modifying the proof for obtaining a result about size is that a circuit $C_{TS}$ only represents the function $f$ on the strings of a given fixed size $n$. The existence of a $C_{SS}$ of size polynomial in that of $C_{TS}$ therefore only allows inverting the function $f$ on the strings of a given size $n$. In other words, if for every $C_{TS}$ there is a polynomially sized $C_{SS}$, then the inverse of $f$ is only polynomial "slice-by-slice": for each restriction of $f$ to strings of a given size $x$ there exists an inverse function that is polynomial-time. This implies that the inverse of a polynomial-time permutation is only *non-uniformly* polynomial-time.

We denote this result and the following ones based on similar proofs with **TS**$\overset{u}{\nrightarrow}$**SS** with the "u" meaning "unlikely". The reducibility results found so far are graphically represented as follows.

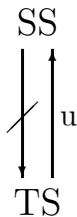

## 4.2 Succinct Sequences of Actions

As for sequences of states, a sequence of actions can be represented by a function from time points to actions, or a function from states to actions, leading to the two following representations.

**time/action representation:** a plan is represented by a triple $\langle s_0, A, C_{TA}, N \rangle$, where $s_0$ is the initial state, $A$ is a set of actions, $C_{TA}$ is a circuit that tells the action to apply at a given time point, and $N$ is the length of the plan;

**next action representation:** a plan is represented by a triple $\langle s_0, A, C_{SA}, N \rangle$, where $s_0$ is the initial state, $A$ is a set of actions, $C_{SA}$ is a circuit that tells the action to apply in any given state, and $N$ is the length of the plan[1].

Together with the two representations of sequences of states, we have four possible ways for representing a sequence. We compare their space efficiency. The following three results are trivial, and therefore not formally stated as theorems.

**SS→SA and SS→TA:** both conversions are based on taking $C_{SS}$ to be the definition of the single action $a$; the sequence of states can be generated by repeatedly applying $a$ regardless of the current state or the time point; the next action and state/action representations of this sequences are made of the circuit that always outputs $a$ regardless of time and current state;

**SA→SS:** a sequence in the next action representation can be polynomially converted into the next state representation: given a state, we can compute the next one polynomially (find the next action, and apply it); therefore, the function that computes the successor of each state is polynomial, and can therefore be represented by a polynomial circuit;

By transitivity, a sequence in SA can be polynomially translated into TA. However, the representations of the sequence in SA and TA may have different sets of actions. This issue will be taken into account in the next section. The results proved so far about reducibilities between representations are summarized by the following figure.

---

[1]This representation can be seen as an implementation of universal planning: a detailed comparison is in the conclusions.

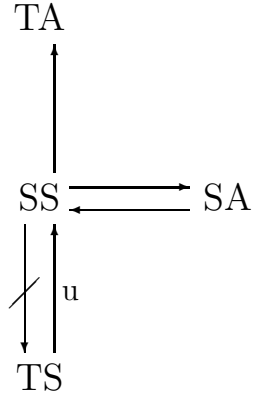

We now show that some plans can be shortly represented in the time/action representation but cannot in the time/state one. This is proved by showing that the time point problem is hard in the time/action representation.

**Theorem 11** *The time point problem is $\|\!\sim\!$NP-hard, if the sequence is in the time/action representation.*

*Proof.* The idea is that some actions can leave some parts of the state unchanged. As a result, we can carry some information from one state to another. This is what makes this representation more powerful of the time/action representation, where the state must be completely determine from the time point.

Let $F$ be a formula of which we want to test satisfiability. We consider a sequence of lenght $2^n$, where $n$ is the number of variables of $F$. Every time point in this sequence can be seen as a propositional interpretation over the variables of $F$. As a result, we can say that $F$ is valid or unvalid in a time point $t$. We only need one state variable, which is initially false.

We only need two actions *sat* and *unsat*, which have very simple effects: *sat* makes the only bit of the state 1, while *unsat* has no effects. The time/action circuit is as follows: given a time point $t$, apply action *sat* or *unsat* according to the validity of $F$ in $t$. At the end of the sequence, the last bit of the state is 1 if and only if the formula is satisfiable.

This proof is correct. However, plans cannot have repetitions of states. Therefore, we modify the state by making it a sequence of $n+1$ bits. The first bit is the satisfiability of the formula as before, while the other $n$ ones are simply copies of the time point: in the initial state they are all zero, and are increased by one by the actions *sat* and *unsat*.

In order to get a proof of $\Vdash$NP-hardness, we use a number of other state variables for encoding all possible 3cnf formulae $F$ of a fixed number of variables. The sequence of states will be then composed of a chunk for each formula, and test the state at the end of the corresponding chunk, as in the proof of Theorem 9. $\square$

As an immediate consequence of the fact that the time point problem is easy in the time/state representation, we have that $\mathbf{TA} \not\to \mathbf{TS}$. This allows adding a new arrow to the figure.

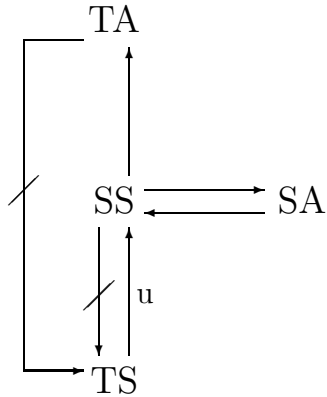

From the previous results, we know that sequences in the next action representation can be polynomially converted into the time/action representation. We now prove that the converse is not possible. To this aim, we show a problem that is easy in the next action representation but hard in the time/action one.

**Definition 6** *The next action problem is that of determining whether an action is the one to execute in a given state.*

This problem only makes sense when the representation used for the sequences include a set of actions. As a result, we can use this problem only for comparing the time/action and the next action representations. However, actions will also be included in the other two representations in the next section.

This problem is clearly polynomial if the plan is in the state/action representation. We have to show that it is hard in the time/action representation.

**Theorem 12** *The next action problem is $\Vdash$NP-hard, if the plan is in the time/action representation.*

*Proof.* The proof is similar to that of Theorem 9. Namely, we have a separate chunk of actions for each possible formula over $n$ variables. The state at the end of any such chunk tells the satisfiability of the formula.

| Formula ($m$ bits) | Model ($r$ bits) | $x$ | $y$ | |
|---|---|---|---|---|
| $F'$ | $00000\ldots00$ | $0/1$ | $1$ | |
| $F$ | $00000\ldots00$ | $0$ | $0$ | |
| | $\vdots$ | | | |
| $F$ | $01101\ldots01$ | $0$ | $0$ | first satisfying model of $F$ |
| $F$ | $01101\ldots10$ | $1$ | $0$ | in the next state, $x$ goes to 1 |
| | $\vdots$ | | | |
| $F$ | $11111\ldots11$ | $1$ | $0$ | |
| $F$ | $00000\ldots00$ | $1$ | $1$ | |
| $F''$ | $00000\ldots00$ | $0$ | $0$ | |

Some changes are needed to make this sequence representable with a time/action circuit. Namely, we are forced to tell exactly which action is executed in each time point. This means that the successor of a state is only determined by whether the action has a given effect in that state or not, and not on which actions is executed in the time point. For example, to make the transition of the $x$ bit from 0 to 1, we cannot use two separate actions to execute in alternative, but we need a single action, whose effect is to set the $x$ bit depending on the satisfiability of the formula.

The problem is that of checking whether an action is to be executed in a state, not to check the occurrence of a state at a given time point. This is however not a problem: when we arrive at the last state of the sequence $F0\ldots0x1$, we execute the action $a$, which begins the new chunk. This means that the action $a$ is executed in $F0\ldots001$ if and only if $F$ is satisfiable. The only remaining point is to ensure that the state $F0\ldots001$ occurs in the sequence, and it has to occur once. This can be done by introducing a new state in the sequence, which is exactly $F0\ldots0(\neg x)1$. Moreover, $a$ has the only effect of inverting the $x$ bit, while the next action starts the new chunk.

In order to ensure that the state occurs in the sequence, we have to define the sequence in such a way the state occurs at $t_1$ if the formula is satisfiable, and at $t_2$ if it is not. If, for example, $t_1 < t_2$, and the state is the $t_1$-th of the sequence, then we can cut the sequence by directly applying the actions of $t_2$. This is still a plan. This means that the sequence of actions we have defined cannot be a minimal plan of any planning instance. □

This theorem implies **TA↛SA**. Together with the previous results, the graph of reducibility becomes as follows.

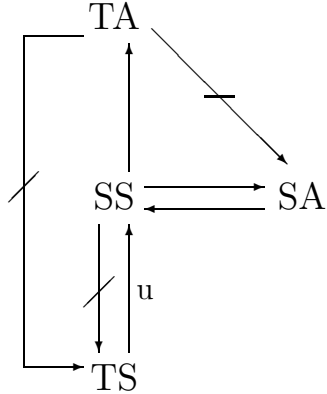

We now consider whether **TS→TA** holds or not. Some evidence indicates that this question has a negative answer. The difficult part of converting $\langle C_{TS}, N \rangle$ into $\langle s_0, A, C_{TA}, N \rangle$ is that both $A$ and $C_{TA}$ must be polynomial in the size of the original representation. Since we can only have a polynomial number of actions, the circuit $C_{TA}$ can only have a logarithmic number of outputs. While $C_{TS}$ is an unrestricted circuit, the combination of $C_{TA}$ and $A$ can be considered as a single circuit composed of two "stages", where only a logarithmic number of variables are shared between the two stages.

If repetition of states were allowed, we could use the simple following sequence of states: the state at an even time point is equal to the time point itself; the state at an odd time point is composed of all zeros. A time/state representation of this sequence is very easy to produce, as the state in every time point can be calculated in a very simple way from the time point. On the other hand, while converting this sequence into the time/action representation we have the problem that only a polynomial number of actions can be generated. Since all states at odd time points are equal, all actions give the same results on them. Since the successors of the odd-time states are all different, that means that we need an exponential number of actions. This proof cannot however be used for sequences of states with no repeated elements. The point is that all odd-time states must be different from each other.

The following theorem shows that $\textbf{TS} \stackrel{u}{\not\to} \textbf{TA}$, that is, the existence of a reduction from the time/state representation into the time/action representation is unlikely.

**Theorem 13** *If every time/state circuit can be converted in polynomial time into an equvalent time/action one, then every polynomial-time permutation is invertible.*

*Proof.* Given an arbitrary polynomial permutation $f$, we can consider sequence composed by $f(x)0$ followed by $x1$, for $x$ ranging from 0 to $2^m$. This sequence is easy to represent in the time/state representation. Let us now assume that there exists $A$ and $C_{TA}$ encoding the same sequence. This implies that the permutation is invertible. Indeed, given $f(x)$, we know that the state that is the successor of $f(x)0$ is $x1$. We however also know that one of the actions in $A$ is executed in $f(x)0$ and results in the next state $x1$. We can therefore determine the state that results from applying each action $a_i \in A$ to the state $f(x)0$. If this is a state $y1$ with $f(y) = f(x)$, we know that $x = y$, as $f$ is a permutation. As a result, we have inverted the permutation $f$ by simply evaluating the result of each circuit in $A$ with input $f(x)0$. Since the time/action representation is polynomial in size, so is $A$: therefore, the algorithm runs in polynomial time, and inverts the polynomial permutation $f$. □

This proof, as the one of Theorem 10, only shows the impossibility of translating from the time/state representation in to the time/action one in polynomial time. The proof can be extended to provide a result about the size of the result of this translation, but weakening the conclusion to the invertibility of permutations in *non-uniform* polynomial time.

As the proof of Theorem 10, this proof is based on a sequence of states for which it is difficult to produce a set of actions for moving from one state to the next one. On the other hand, sequences that are plans of planning problems results from problems in which a set of actions is already specified. This issue is considered in the next section. The following figure shows the results obtained so far.

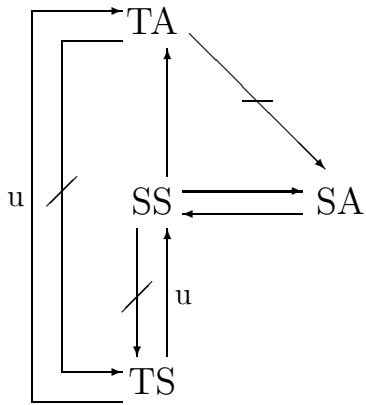

Remarkably, all arrows from TS are labeled with "u". In other words, the non-reducibility of this representation to the other ones is conditioned to non-uniform non-invertibility of polynomial permutations.

## 4.3 Planning

In this section, we analyze the problem of planning. An instance of the problem of planning is composed of an initial state, a specification of the actions, and one or more goal states. In this paper we consider a language that is general enough to include languages such as STRIPS and $\mathcal{A}$, while retaining the property that the result of executing an action in a state can be determined in polynomial time.

**Definition 7** *A POLYPLAN instance is a quadruple composed of: a set of $n$ boolean variables, two propositional interpretations on them (the initial and the goal state), and a set of actions; each action is a circuit with $n$ inputs and $n$ outputs, telling the next state from the previous one.*

A plan is usually defined as a sequence of actions from the initial state to the goal state. However, a plan can also be defined as a sequence of actions whose first and last states are the initial and the goal states, respectively. Given a sequence of actions, determining the corresponding sequence of states can be simply done by iteratively evaluating the result of the actions starting from the initial state. A sequence of actions thus corresponds to a single sequence of states. On the other hand, a sequence of states can be generated by zero, one, or more sequences of actions. However, determining such sequences is easy: for each pair of consecutive states, evaluate the result of all actions on the first state and check whether the result is the second state. As a result, one can translate a sequence of states into a sequence of actions in polynomial time, and vice versa.

Sequences of actions and sequences of states are instead not equivalent when a succinct representation is used. Moreover, both sequences of actions and sequences of states can be represented into two different succinct ways, differently from QBF, in which there is a single natural representation of a model. We therefore have a total of four different representations of plans. In the previous section, we have compared the succinct representations of sequences. We however have already noticed that not all sequences and not all results are reasonable in the scenario of planning, in which a fixed set of actions is given. Using a fixed set of actions makes some results of the previous section invalid. We consider this issue in the next section.

### 4.3.1 Succinct Sequences that Are Plans

The succinctness analysis of representations of sequences does not take into account that sequences that are plans for a given instance of the planning problem have some properties. In particular, if the actions of the instance are $A$, sequences of states and sequences of actions must obey the two following conditions:

**Sequence of States:** for each pair of consecutive states $s_i$ and $s_{i+1}$ the set $A$ contains an action such that $s_{i+1}$ is the result of applying the action to $s_i$;

**Sequence of Actions:** the set of actions in both representations $\langle s_0, A, C_{SA}, N \rangle$ and $\langle s_0, A, C_{TA}, N \rangle$ must coincide with the set of actions of the planning instance.

A consequence of the second point is that the two translations **SA→SS** and **SS→TA** do not necessarily imply that **SA→TA**. Indeed, the set of actions of the initial SA representation is lost while translating the sequence into SS (which is a sequence of states). While translating SS into TA, a new set of actions is generated, which can be different from the original one. This issue can be solved by adding the set of actions also to the two representations of sequences of states. A translation is then considered valid only if the set actions is not changed.

Adding the actions to the representations of sequences of states has the additional benefit that a sequence of states can be validated against the actions: we can verify whether the transition from a state to the next one actually corresponds to one of the actions.

This addition also forbids the use of sequences that are exponentially shorter than the original planning instance in the proofs of non-reducibility. Consider the proofs based on the existence of one-way permutations. These proofs are based on sequences in which a state $f(x)0$ is followed by $x1$, and determining $x$ from $f(x)$ is assumed not possible in polynomial time. This means that the transition from the state $f(x)0$ to $x1$ cannot be done by a set of actions unless this set takes space exponential in the size of the state variables. Which implies that the TS representation of this sequence is exponentially shorter than the original planning instance. By adding the set of actions to this representation, such a sequence is made as large as the original planning instances, and the proof of non-reducibility does not hold any more.

Summarizing, we only want to consider the sequences of states or actions that results from a single given instance of planning, and this can be done as follows:

1. All representations for sequences contain the initial state and the set of actions;

2. a valid representation for a sequence should be such that there exists an action leading from each state to the next one;

3. we only consider reductions with the same set of actions.

The representations of sets of actions TA and SA are not affected. The representations of sets of states TS and SS must be changed by adding $s_0$ and $A$ to them. These two represententations therefore become $\langle s_0, A, C_{TS}, N \rangle$ and $\langle s_0, A, C_{SS}, N \rangle$. Moreover, for any two consecutive states $s_i, s_{i+1}$ of the sequence, the set $A$ must contain an action $a$ such that $s_{i+1}$ is the result of performing $a$ on the state $s_i$.

For sequences of states, the actions are not really needed to represent the sequence. However, they are considered as a part of the representation to consider the size of the domain into account: indeed, if a sequence is only generable by an exponentially-size domain description, then adding $A$ to the sequence representation makes it exponential as well. This prevents from calling "polynomially representable" a sequence which is actually only generated by an exponential number of actions or by exponentially sized actions. Such sequences indeed exists (the sequence in which $f(x)0$ is followed by $x1$ is one such sequence.)

Let us now consider the reductions we have already proved:

**SA→SS:** given an SA representation, we can compute the successor of each state in polynomial time by first computing the action and then its result; therefore, a polynomially sized circuit $C_{SS}$ exists representing this function; the rest of the SS representation is the same as the SA original one;

**SS→SA:** we are given the representation $\langle s_0, A, C_{SS}, N\rangle$, and want to determine a circuit that tells the action to execute in each state. To determine the action to execute in $s$, we can simply evaluate $C_{SS}$, and select the first action that leads from $s$ to it. This can be done polynomially; therefore, a polynomial circuit telling the next action exists;

**SS→TA:** here, the reduction was based on the fact that we can always encode the $C_{SS}$ circuit into a single action, which is then applied in all time points. However, this is no more the case now: we can have more than one action already in the SS representation, and therefore need to know which one to apply in a time point.

As a result, the only reduction that is no more valid is **SS→TA**. We consider the equivalent reduction **SA→TA**, and prove it impossible under the assumption that the set of actions is the same. In other words, while the translation is possible by altering the set $A$, it is impossible if both representations contain the same set of actions. Let us therefore consider the following problem.

**Definition 8** *The time/action problem is that of determining the action to execute in a given time point.*

This problem is easy if the sequence is in the time/action representation, as it amounts to determine the output of the circuit $C_{TA}$, given the time point as input. We can prove that this problem is however hard if the sequence is in the state/action representation.

**Theorem 14** *The time/action problem is $\|\hspace{-0.3em}\leadsto$NP-hard, if the sequence is in the state/action representation.*

*Proof.* We use the usual sequence encoding the satisfiability of a formula. Each state is composed by a formula $F$, a model $M$, and another bit $x$. The sequence contains a chunk for each formula; in turns, each chunk contains a state for each model.

We only have four actions *ok*, *no*, *sat*, and *unsat*. The effects of these actions are as follows:

**ok** the succcessor of a state $FMx$ is the state $F(M+1)1$, if $M$ satisfies $F$; otherwise, the state is not changed;

**no** the succcessor of a state $FMx$ is the state $F(M+1)x$ if $M$ does not satisfy $F$; otherwise, the state is not changed;

**sat** the successor of a state $FM_f1$ is the state $(F+1)M_i0$, where $M_i$ and $M_f$ are the first and last interpretation, respectively; the succeessor of $FM_f0$ is $(F+1)M_i1$ if $M_f$ satisfies $F$; for all other cases, the next state is the same as the current one;

**unsat** the successor of a state $FM_f0$ is the state $(F+1)M_i0$, where $M_i$ and $M_f$ are the first and last interpretation, respectively, if $M_f$ does not satisfy $F$; for all other cases, the next state is the same as the current one;

The sequence is that obtained by scanning each model for each formula. The actions that realize this sequence are: for each state $FMx$, in which $M$ is not $M_f$, we apply either *no* or *ok*, depending on whether $M$ satisfies $F$ or not; in the state $FM_fx$ we apply either *sat* or *unsat*.

The reduction simply works because the action to execute at the end of the chunk corresponding to a formula $F$ is *sat* or *unsat* depending on the satisfiability of $F$. $\square$

This theorem proves that $\mathbf{SA} \not\Rightarrow \mathbf{TA}$. The proof that $\mathbf{TA} \not\Rightarrow \mathbf{SA}$ still holds if we enforce the actions to be the same: indeed, the result is proved by showing that the problem of the next action is easy in the representation SA but hard in TA: however, the problem only makes sense for sequences represented by the same set of actions. The following figure illustrates the results proved so far.

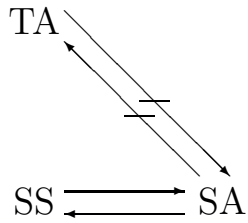

Let us now consider the reduction from and to the time/state representation. The fact that **SS↛TS** can still be proved to hold: the proof of hardness of the time point problem when the sequence is in the SS representation is based on a sequence that can be generated by a single action. The result **TA↛TS** still holds because the proof of hardness of the time/state problem in the TA representation clearly involves sequences whose representation include the actions.

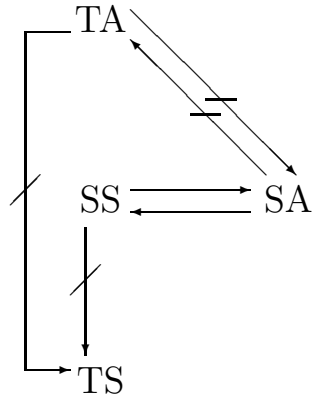

Let us now consider the proofs of non-reducibility from TS to SS and TA. Both proofs are based on sequences in which $f(x)0$ is followed by $x1$, where $f$ is a polynomial permutation. However, if the representation of a sequence includes the actions that allows making such transitions, then such actions have to be exponential (in size or number), otherwise all polynomial permutations are invertible.

On the contrary, the reduction from TS to TA is always possible: given a time/state circuit, we can determine the corresponding time/action one:

given a time point $t$, the action to execute is simply one that lead from the state $s_t$ to $s_{t+1}$, where these two states are determined from the time/state circuit. Such an action exists by assumption.

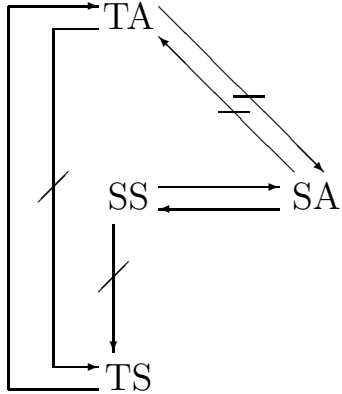

We link the existence of a reduction from TS to SS with the existence of one-way polynomial functions.

**Theorem 15** *If translating from a time/state representation into a next state representation with the same actions can be done in polynomial time, then every injective polynomial-time function is invertible.*

*Proof.* Let $f$ be a polynomial-time injective function. Assume that, for every $x$ of size $n$, the lenght of $f(x)$ is exactly $n$: this can be achived by padding both $x$ and $f(x)$ with blanks (we are not assuming that $f$ is a permutation.)

We show that a transition from the time/state representation to the next state one implies the invertibility of $f$. For every $x$, consider the following chunk:

$$\begin{array}{ccc} f(x) & f(x) & 0\ldots 0 \\ & \vdots & \\ f(x) & x & 1\ldots 1 \\ f(x+1) & f(x+1) & 0\ldots 0 \end{array}$$

Since no polynomial action can turn $f(x)$ into $x$, we add $n$ intermediate states between them. In each time step we only change two bits: a bit of $f(x)$ is turned into a bit of $x$ and the corresponding bit of $0\ldots 0$ is turned into 1. For this to happen, we only need $2n$ actions: the $i$-th and the $n+i$-th actions chang the $n+i$-th bit of the state to 0 and to 1, respectively, and both change the $2n+i$-th bit into 1.

The idea is that going from $f(x)$ to $x$ is done by a number of very simple actions. What is difficult in going from $f(x)$ to $x$ is the choice of actions to apply. A single action is then necessary to turn $f(x)x1\ldots 1$ into $f(x+1)f(x+1)0\ldots 0$, as going from $x$ to $x+1$ to $f(x+1)$ is by definition polynomial.

This sequence contains no repetitions because each chunk contains an unique prefix $f(x)$ and each state of the chunk has a different value in the last $n$ bits. Given a time point $t$, the first $n$ bits of it are equal to $x$. Since $x$ allows determining $f(x)$ in polynomial time, all states of the chunk can be determined in polynomial time from the time point. As a result, a time/state representation of this sequence of size polynomial in $n$ exists.

Let us now assume that this representation can be translated into a next state representation with the same actions in polynomial time. Then, given $f(x)$ one can produce the time/state representation of the above sequence, translated it into the next state representation, and calculate $x$ by evaluating the next state circuit $n$ times from the state state $f(x)f(x)0\ldots 0$. This results in a state containing $x$. Therefore, every polynomial-time injective function is invertible. □

As for Theorem 10, this result can be extended to the size of the result of the translation, but the conclusion of the claim would only be that the inverse of injective polynomial-time functions are *non-uniformly* polynomial. The relationship between representations are summarized in the following figure.

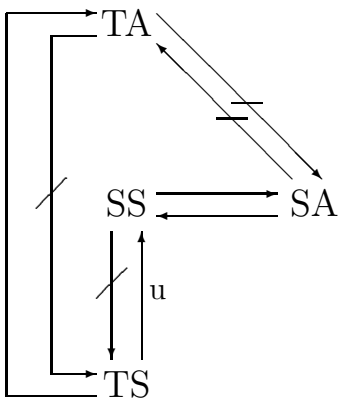

### 4.3.2 Bounding Matters

The problem of compactly representing the models is easy for QBFs, as there is only one natural succinct representation for models. On the contrary, at least four different succinct representations for sequences are possible, and this complicates the analysis of planning.

Having decided how to compactly represent the solutions of POLYPLAN, we can now turn to the second step of the program of analysis, that is, proving that the problem is affected by bounds on the size of the solution. This step is really straightforward if we consider a fixed $k$ as a bound: the problem of finding a sequence, and the problem of finding a sequence representable in space bounded by $k$ are different problems. It is indeed easy to show instances of planning having only plans of size two or more, and that cannot be stored (even in succinct representation), in space less than $k = 1$. This result is trivial; what is more interesting to prove is that the same happens for any fixed $k$.

**Theorem 16** *For every $k$ there exists a POLYPLAN instance, of size polynomial in $k$, that has plans that cannot be represented in space $k$, using any of the four representations for sequences.*

*Proof.* The specific proof depends on the kind of representation we consider, but the main idea is the same: we show a POLYPLAN instance whose plans contain more than $k + 1$ actions. Such sequences can be represented with a circuit having less than $k + 1$ gates only if some output line is directly connected to some input line. We therefore use a sequence of actions that does not allow such a direct connection. This is the part that depends on the representation. Once this is proved, it shows that any circuit representing such plans require at least $k + 1$ gates. □

The proof is based on a POLYPLAN instance whose plans are longer than $k$ steps. The result then follows from the fact that there are sequences longer than $k$ that cannot be represented with $k$ bits. The proof is only made complicated by the fact that we are restricting to sequences that are plans of POLYPLAN instances.

A more interesting question is to check whether any POLYPLAN instance that has plans, has also plans that are representable in polynomial space in some succinct form. This is the Polynomial Solution Property (Property 1) applied to planning.

This property holds or not depending on how plans are represented. For example, if plans are in non-succinct form, then there are instances of POLYPLAN whose plans are all exponentially long; since plan length coincide with plan size for the non-succinct representation, the polynomial plan property does not hold.

For some other representations the property holds. Namely, it holds for the following, quite artificial, representation of sequences: a sequence is represented by a triple composed by a. the POLYPLAN instance $P$, b. an ordering of the actions $(a_1, \ldots, a_m)$, and c. an integer number $N$. The sequence represented in this way is simply the $N$-th plan of $P$, where plans are ordered lexicographically according to the order on actions. This is a quite awkward representation for sequences: still, it is a way for representing them.

Using this representation, any instance that has plans has also the plan in which the integer $N$ is 1 (this is simply the first plan, in the lexicographic order.) In our ugly representation, this sequence takes only a linear amount of space (the space for representing $P$, the ordering of actions, and the number $N = 1$.) We can therefore say that the polynomial plan property holds for this representation: any instance having solutions also have solutions that can be represented in polynomial space.S

Since the polynomial property does not hold for the explicit representation but holds for the artificial succinct representation above, its validity depends on how sequences are represented.

We prove that the polynomial plan property does not hold for the four succinct representations introduced in the previous sections. The proof is based on exhibiting a problem that is easy to solve given a plan in a succinct representation but hard given the original planning instance.

A difficulty of this strategy is that a succinct representation of a plan represents a single plan, while a planning instance may have more than one plan. For example, a problem like: "is there a plan starting with the action $a$" is ill-posed if only a sequence is given: from the sequence, we can determine the first action, but we cannot say anything about the actions of the other plans.

To avoid this problem, we only consider planning instance that have a single plan. This way, every problem that can be defined on a plan can equivalently defined on the planning problem that has only that single plan. A planning instance having a single plan can be considered a representation of a sequence.

**unique plan representation:** a sequence of states or actions is represented by a POLYPLAN instance having this sequence as its only plan.

This representation is clearly not useful in practice, but is a useful device for proving that some planning instances have plans, but none of them can be represented in polynomial space with the other succinct representations. Namely, we show that some sequences that can be represented in polynomial space using the UP representation but cannot in the other ones. This way, proving that bounding alters the problem can be done with the technique used to compare plan representations: we show that some problems are hard in the UP representation but easy in the other ones. The following problems are easy in the succinct representations:

**time/state problem:** easy for the time/state representation;

**next state problem:** easy for the next state representation;

**time/action problem:** easy for the time/action representation;

**next action problem:** easy for the next action representation.

For example, the next state problem can be solved easily, if we know a next state representation of the sequence. If the same problem is $\|\hspace{-2pt}\leadsto$NP-hard for the UP representation, then the conversion from UP to SS cannot always be done polynomially. In less formal terms, there are instances of planning having a single plan that cannot be represented in the next state representation in polynomial space. We indeed prove that all problems in the list above are hard for sequences in the unique plan representation.

**Theorem 17** *Given a POLYPLAN instance that has a single plan, the problems of the next state, next action, time/state, and time/action are all $\|\hspace{-2pt}\leadsto$NP-hard.*

*Proof.* Consider the sequence composed by a chunk for each formula $F$ on an alphabet of $n$ variables, each chunk being made as follows:

$$
\begin{array}{ccccc}
0 & 0 & 0 & F & M_0 \\
1 & 0/1 & 0 & F & M_1 \\
1 & 0/1 & b & F & M_1 & b = 1 \text{ if } M_0 \text{ satisfies } F \\
1 & 0/1 & c & F & M_2 & c = 1 \text{ if } M_0 \text{ satisfies } F \\
& & \vdots & & \\
1 & 0/1 & 0/1 & F & M_f
\end{array}
$$

The first bit is changed in the first step. The second bit may change to 1 in the first step, but maintains its value from this point on. The third bit is changed only if the formula $F$ is satisfied by the model $M_i$; otherwise, it maintains its value.

We only need three actions: *sat*, *unsat*, and *move*. The first two actions can only be executed if the first bit is 0, and change it to 1. Therefore, they can only be executed in the first state of each chunk. The *sat* action also changes the second bit to 1, while *unsat* does no other change.

The *move* action can be used only if the first bit is 1, and does two things: first, it moves from a model to the next one; second, it sets the third bit of the state if the model satisfies the formula. In the last state of the sequence, *move* leads to the first state of the next chunk, but only if either of the three following conditions are true:

1. the second and the third bits are equal to 0 and $M_f$ does not satisfy $F$;

2. the second and the third bits are equal to 1,

3. the second bit is 1 and $M_f$ satisfies $F$.

By setting the initial state to be the first state of the first chunk, and the final state to be the last state of the last chunk, these actions can be arranged into a plan only if we travel all chunks in order. This is in turns possible only if the first action is either *sat* or *unsat*, and we then repeat the action *move* several times. At the end of the chunk, the action *move* actually moves to the next chunk, but only if the first action of the chunk (*sat* or *unsat*) reflects the satisfiability of the formula. Indeed, if we perform *sat* at the beginning of the chunk, then the second bit is set to 1; the *move* actions then sets the third bit to 1 if and only if the formula is satisfiable, and this is the only condition allowing to move to the next chunk. In the same way, if we first execute *unsat* then we can move to the next chunk at the end of this one only if the formula is unsatisfiable.

This way, not only the POLYPLAN instance has exactly one plan, but this plan encodes the satisfiability of all formulae of $n$ variables. Indeed, the action to execute in the state $000FM_0$ is *sat* or *unsat* if and only if the formula $F$ is satisfiable. This proves that the problem of the next action is $\|\rightsquigarrow$NP-hard.

The problem of the next state is hard for the same reason: the successor of $000FM_0$ is $110FM_0$ if and only if $F$ is satisfiable. Finally, the time/state problem and the time/action problems are $\|\leadsto$NP-hard as well because all chunks have the same length: given a formula, we can check its satisfiability by checking the next state or the next action from the initial state of the corresponding chunk. □

As a consequence, the polynomial plan property does not hold for any of the four succinct representations.

**Corollary 3** *If the polynomial size property holds for any of the four succinct representation, then* NP $\subseteq \|\leadsto$P.

### 4.3.3 Complexity: Unary Notation

**The Time/State Representation** The problem of checking the existence of a plan with a time/state representation of size bounded by a number $k$ in unary can be expressed as follows: there exists a circuit $C_{TS}$ and an integer $N$ such that: the initial state is $C_{TS}(0)$; the last state $C_{TS}(N)$ satisfies the goal; from each state $C_{TS}(t)$ to the next one $C_{TS}(t+1)$ there is an action; and no state occurs twice in the sequence.

$$\begin{aligned}\exists C_{TS}, N \ . \quad & C_{TS}(0) = s_0 \\ & C_{TS}(N) \in GOAL \\ & \forall t, \exists a \in A \ . \ C_{TS}(t+1) = RESULT(a, C_{TS}(t)) \\ & \forall t, t' \ . \ t, t' < N, t \neq t' \Rightarrow C_{TS}(t) \neq C_{TS}(t')\end{aligned}$$

The quantification over the actions can be converted into a disjunction, as there are only polynomially many actions. The only two left quantifiers are the existential quantifier over circuits and the universal quantifier over time points. Since $k$ is in unary, the first quantification can be expressed as a QBF with polynomially many bits. The problem is therefore in $\Sigma_2^p$.

Proving that the problem is NP-hard is easy: POLYPLAN is NP-hard even if the length of the searched plan is restricted to be polynomial in the instance size. In this case, existence of a plan coincide with existence of a succinct plan of size bounded by the same polynomial. We can also prove that the problem is $\Sigma_2^p$-hard, and therefore complete for that class.

**Theorem 18** *Deciding whether a* POLYPLAN *instance has a plan that can be represented in space $k$ with a time/state circuit is $\Sigma_2^p$ complete, when $k$ is represented in unary notation.*

*Proof.* Membership is shown above. Hardness is proved by reduction from $\exists\forall$QBF: given a formula $F = \exists X \forall Y.E$, we build the following POLYPLAN instance. States are propositional interpretations over the alphabet $\{g\} \cup Z \cup X \cup Y$, where $g$ is new variable and $Z$ a set of new variables in one-to-one correspondence with those of $X$. The initial state $s_0$ sets all variables to false. The goal state $s_g$ is the one setting the first variable to true and the other ones to false. The actions are the following ones:

*settrue*: can only be executed when some bits of $Z$ are false; namely, it sets to true the first $z_i$ that is false, and then sets the corresponding $x_i$;

*setfalse*: can only be executed if some bits of $Z$ are false; it sets the first zero bit of $Z$ to true;

*move*: possible only if all bits of $Z$ are true and some bit of $Y$ is false; it increases $Y$ by 1, but only if the values of $X$ and $Y$ satisfy the formula $E$;

*last*: possible only if all bits of $Y$ are true; it goes into the goal state, but only if the value of $X$ and $Y$ satisfy $E$.

Any plan of this instance must be composed of two parts. In the first part, we execute $n$ times an action between *settrue* and *setfalse*. The result of this phase is a state in which all variables $Z$ are true, and $X$ is a model that depends on which among *settrue* or *setfalse* we have executed. From this point on, we can only execute *move* to increase the model $Y$ from the first to the last one. This movement is however only possible if all values of $Y$, together with the value of $X$ (which does not change any more) satisfy $E$. Finally, the last action is *last*, leading to the goal state only if the last model of $Y$ satisfies $E$ as well.

This sequence of actions is a plan if and only if the value of $X$ that results from the first phase makes $E$ satisfied for all possible values of $Y$. The instance has a plan if and only if there is a sequence of actions in *settrue* and *setfalse* that leads to an $X$ with this property. Therefore, the existence of plans is equivalent to the existence of a value of $X$ such that all models over $Y$ satisfy $E$. □

This theorem provides an alternative proof of the fact that bounding affects the problem of checking the existence of plans. Indeed, the problem without bounds is PSPACE-hard, while the bound makes it in $\Sigma_2^p$. As a result, the problems are different unless PSPACE=$\Sigma_2^p$.

**The Other Succinct Representations** The problem of checking the existence of a plan for the other succinct representations can be shown to be PSPACE complete.

**Theorem 19** *Deciding whether a* POLYPLAN *instance has a plan that can be represented in space k with the next state, next action, or time/action representation, is* PSPACE *complete, when k is in unary notation.*

*Proof.* Membership follows from the fact that *checking* whether a sequence is a plan can be done in polynomial space: given any representation of a sequence, we can start from the initial state, and then generate one-by-one all the other states of the sequence. Since we can forget a state once the next is known, this algorithm only takes a polynomial amount of memory. Checking the existence of a plan of bounded size can be done by guessing a sequence and then checking whether it is actually a plan. Since the verification step is in PSPACE, the whole problem is in PSPACE as well since PSPACE is equal to nondeterministic-PSPACE.

Hardness is proved by encoding a generic Turing machine working with a polynomial amount of memory into the problem of determining the last state of a sequence in any of the three representations. This is done as follows: the state are the possible states of the memory; there is only one action, which acts like the Turing machine in modifying a state into the next one (note that determining the move of a Turing machine is very easy.) The initial and the goal state are the initial and the goal state of the Turing machine.

Since the instance only contains one action, planning is equivalent to verifying that the goal state can be reached from the initial state by repeating the executing of the action.

To complete the proof, we only have to verify that planning and planning with bounds are the same in this case. We actually prove that the plan of this instance (if any) is representable in very little space in all three representations. For the time/action and the next action representation this is easy: we simply execute the same action regardless of the time point or current state. The next state representation of this plan is also easy to find, as the circuit $C_{SS}$ that gives the next state is simply equal to the only action of the instance.

We can therefore conclude that checking the existence of plans that can be represented in space $k$ is PSPACE complete, if $k$ is represented in unary and the plan is represented using any of the following representations: next state, time/action, and next action. □

Note that we cannot use this theorem for proving that not all instances having plans also have polynomially-sized plans as we did for the time/state representation.

### 4.3.4 Complexity: Binary Notation

In the case of binary representation of $k$, the problem is PSPACE-hard regardless of the representation of plans. This is a straightforward consequence of the fact that we can use an exponentially large value of $k$, and we can then use a value that allow any possible plan.

**Theorem 20** *The problem of existence of plans of size bounded by an integer $k$ in binary representation is* PSPACE-*hard for all four succinct representations introduced in this paper.*

Another result that is independent to the representation is that, if the problem is EXPTIME-hard, or it can otherwise be proved not to be in P, then $P \neq PSPACE$.

**Theorem 21** *If the problem of existence of plans of size bounded by $k$, represented in binary, is* EXPTIME-*hard, then* $P \neq PSPACE$, *for all four succinct representations introduced in this paper.*

*Proof.* We prove that P = PSPACE implies that the problem is in P. We therefore assume P = PSPACE.

Since the problem of existence of a plan is in PSPACE, finding the first action of a plan is in PSPACE as well. Indeed, the first action of a plan can be determined by considering each action in turn, and checking whether the result of executing this action is a planning instance having a plan. The first action for which this is true is the first action of a plan. This procedure can be performed in polynomial space: there is only a linear number of actions, and checking each action is in PSPACE. The whole procedure is therefore in PSPACE.

Once the first action is determined, we can determine the second action of a plan in the same way. Therefore, the whole plan can be produced using only a polynomial amount of space. By checking the action to perform always in the same order, we can say that producing the $n$-th action of this plan can be done in polynomial space. In the same way, producing the state at the $n$-th step of the plan is in polynomial space as well.

Since PSPACE = P by assumption, the problem of generating an arbitrary action or state of this plan is in P. As a result, if PSPACE = P, then a planning instance has a plan if and only if it has a plan that can be represented in any of the four representations in polynomial space.

This means that checking the existence of a plan that is representable in space $k$ is polynomial. Indeed, we know a polynomial $p$ such that an instance has a plan if and only if it has a plan of size bounded by $p$. Checking the existence of a plan bounded by $k$ amounts to checking the existence of a plan whose size is less than the minimal between $k$ and the value of this polynomial. This can be done by a guess-and-check algorithm, and the plan to be guessed can be represented in polynomial space. This proves that the problem is in NP, and therefore it is in P. □

Whether this problem is in EXPTIME or not is an open question.

## 4.4 Linear Temporal Logic

A Kripke models is composed of a set of possible *worlds*, an accessibility relation between words, and a function from worlds to propositional interpretations over the set of variables. This definition can be simplified for LTL by defining a model to be an infinite sequence of propositional interpretations.

Since a model is a sequence of propositional interpretations, we can use the same terminology of planning and define **a state to be a truth interpretation over the variables.** Planning and LTL have therefore in common that a sequence of states is a solution. They are however different because:

1. there is no concept of "action" in LTL;

2. a model is an infinite sequence of states in LTL;

3. the same state may occur more than once in an LTL model.

The first point implies that the next action and the time/action representations of solutions do not make sense in LTL.

The second point implies that not all models of an LTL formula can be represented finitely. In particular, some models of a formula can be represented finitely while some other do not. For example, the models of $\mathbf{GF}x$ are the sequences in which a state satisfying $x$ occurs infinitely often. A finitely representable model is that in which the same state in which $x$ is true repeated infinitely. On the other hand, some models of $\mathbf{GF}x$ cannot be represented finitely. Indeed, a model of $\mathbf{GF}x$ is a sequence composed of an arbitrarily long sequence in which $x$ is false followed by a state in which $x$ is true followed by another sequence in which $x$ is false, etc. These models are in correspondence with the infinite sequences of integers (the distances between states in which $x$ is true). If all such models can be finitely represented, than all infinitely long sequences of integers can be finitely represented as well.

We can however restrict to models that are ultimately periodic. Such models are composed of a first sequence of states followed by a second one which keeps repeating itself. Every satisfiable formula has an ultimately periodic model in which both parts (the initial sequence and the period) are at most exponentially longer than the formula [SE85]. An ultimately periodic model can be represented by a pair of sequences of states, representing the initial part of the sequence and its period. Given a sequence of states, we consider the shortest possible initial part and period, so that the sequence $abcbcbcbc\ldots$ is composed of the initial part/period pair $\langle a, bc \rangle$ rather than $\langle ab, cb \rangle$ or $\langle a, bcbc \rangle$.

The third point above is that the same state may occur more than once in the same LTL model. In particular, the same state may occur more than once even inside the initial part or period of a model. For example, all models of $\neg a \wedge \mathbf{X} \neg a \wedge \mathbf{XX}a \wedge \mathbf{XXX}a$ begins with the the state in which $a$ is false repeated twice followed by the state in which $a$ is true repeated twice. As a result, either the initial part or the period of the models of this formula contains the same state twice. The models of this formula cannot be represented at all in the next state representation. We however still consider the next state representation because some LTL formulae have models that can all be represented in the next state representation and are even more succinct in this representation rather than in the time/state one.

The problems we consider are the following ones.

1. comparison of the succinctness of representations of sequences when they represent models of LTL formulae;

2. bounding matters: a bound on the size of the models may change the satisfiability of a formula;

3. complexity of satisfiability when the size of the succinctly represented models are in unary and in binary notation.

In the rest of this section, we show some general results about LTL. We define $\mathbf{X}^i$ to be the repetition of $\mathbf{X}$ for $i$ times. Formally, for any formula $F$, we define $\mathbf{X}^0 F = F$ and $\mathbf{X}^i F = \mathbf{X}(\mathbf{X}^{i-1} F)$ for any $i > 0$; informally:

$$\mathbf{X}^i F = \underbrace{\mathbf{X} \cdots \mathbf{X}}_{i} F$$

An LTL model is lexicographic over the variables $X$ if the values of the $X$ in the states represent the integers in order. In other words, when the values of $X = \{x_1, \ldots, x_n\}$ are interpreted as numbers of $n$ bits, the value of $X$ in a state is exactly one more than the value in the previous state, where increase is modulo $2^n$. The following is a graphical representation of a lexicographic model of two variables $X = \{x_1, x_2\}$.

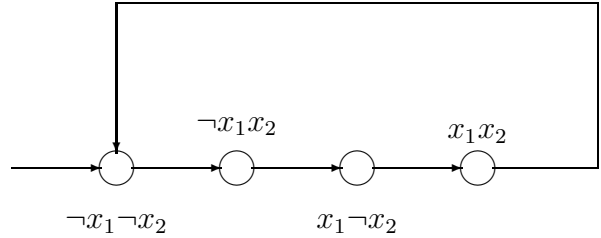

More than one model can be lexicographic over $X$. Indeed:

1. the initial state can be any of the four represented ones;

2. the variables other than $X = \{x_1, x_2\}$ can assume arbitrary values.

Formally, an accessibility relation $R$ is lexicographic on a subset of variables $X$ composed of $n$ variables if $\omega R \omega'$ holds only if the integer represented by the assignment over $X$ of $\omega'$ is exactly the one represented by $\omega$ plus one modulo $2^n$. A model is lexicographic over $X$ if its accessibility relation is lexicographic over $X$.

**Definition 9** *A model is lexicographic over variables $X$ if its accessibility relation $R$ is such that $sRs'$ holds if and only if $\omega' = (\omega + 1) \bmod 2^{|X|}$, where $\omega$ and $\omega'$ are the values of $X$, seen as integers, in $s$ and $s'$, respectively.*

The following lemma tells that there exists a simple LTL formula that enforces its models to be lexicographic over $X$. This is obtained by specifying how the values of $X$ have to change at each step.

**Lemma 2** *All models of $COUNT_X$ are lexicographic over $X$.*

$$COUNT_X = \mathbf{G}(x_n \not\equiv \mathbf{X}\neg x_n) \wedge$$
$$\bigwedge_{i=1}^{n-1} \mathbf{G}\Big((x_{i+1} \wedge \mathbf{X}\neg x_{i+1}) \equiv (x_i \not\equiv \mathbf{X}\neg x_i)\Big)$$

*Proof.* The first subformula $\mathbf{G}(x_n \not\equiv \mathbf{X}\neg x_n)$ forces $x_n$ to change at each step; indeed, this formula is only verified if the value of $x_n$ in each state is different from the one in the next state. The second subformula $\mathbf{G}((x_{i+1} \wedge \mathbf{X}\neg x_{i+1}) \equiv (x_i \not\equiv \mathbf{X}\neg x_i))$ forces $x_i$ to change if and only if the next bit $x_{i+1}$ is true and will be false in the next state. $COUNT_X$ is therefore only satisfied by the models in which the value of the variables in $X$ represent a number that increase by one modulo $2^{|X|}$ at each step. □

The formula $COUNT_X$ only constraints the values of the $X$ to increase at each step, but does not constraint the initial values of $X$ nor the values of other variables besides $X$. This can however be achieved by conjoining $COUNT_X$ with other formulae.

### 4.4.1 Succinct LTL Models

Since an LTL model is a sequence of states, the two representations of sequences of actions are not considered. The next state representation of sequences cannot always be used even if we restrict to ultimately periodic models because a state may appear twice even inside the initial part or the period of a model. Therefore, the time/state representation is in absolute more powerful when considering sequences that are models of LTL formulae. We however consider the relationship between the next state and the time/state representation for models that can be represented in both ways, and prove that the next state representation can be more succinct.

The relationship between the next state and the time/state representations changes if one consider arbitrary sequences or sequences that are plans of some planning domain. This is not the case for LTL. Indeed, every infinite sequence of states is a model of the LTL formula true. Therefore, there is no difference between comparing "sequences" and "sequences that are models of some LTL formula".

However, this argument only shows the different succinctness of the two representations on *some* models of an LTL formula. For example, Theorem 9 proves that there is a model of the LTL formulae true that is more succinct in one representation than in the other one, but the same LTL formula true also have models that are short in both representations.

We therefore study the existence of LTL formulae whose models are all short in one representation but long in the other one. This is done by using LTL formulas having a single model. In particular, we show that every sequence in the next state or in the time/state representation can be modified to become the only models of an LTL formula. The following definitions characterize these translations.

**Definition 10 (Expansion of a Sequence)** *A sequence of states $S_2$ expands the sequence $S_1$, denoted by $S_1 \preceq S_2$, if and only if the states of $S_2$ coincide with the states of $S_1$ on the variables of $S_1$ and all other variables of $S_2$ are always false.*

In other words, the states of $S_2$ are the same as those of $S_1$ but for some other variables that are always false.

**Definition 11 (Division of a Sequence)** *The sequence $S$ divided by an integer $t$, denoted by $S/t$, is the sequence obtained by taking only the elements of $S$ that are in a position that is a multiple of $t$ plus $1$.*

The following two theorems prove that a sequence in one of the two representations can be converted in polynomial time into another sequence in the same representation that is the only model of an LTL formula. The resulting sequence $S_2$ is obtained by adding new variables and new states to the original sequence $S_1$ in such a way $S_1 \preceq (S_2/2)$, i.e., the sequences are as follows:

$$\begin{array}{rccccccc} S_1 & = & & s_1 & & s_2 & & s_3 & \ldots \\ S_2 & = & s'_1 & s_1 s_{\mathsf{false}} & s'_2 & s_2 s_{\mathsf{false}} & s'_3 & s_3 s_{\mathsf{false}} & \ldots \end{array}$$

In words, every state $s_1$ is added a part that evaluates all other variables to false. Before the first state and between each pair of states there is a new state.

**Theorem 22** *Every sequence of states $S_1$ in the next state representation can be translated in polynomial time into an LTL formula $L$ and another sequence $S_2$ in the next state representation that is the only model of $L$ and $S_1 \preceq (S_2/2)$.*

*Proof.* Let $\langle s_0, C_{SS}, N \rangle$ be the next state representation of the sequence $S_1$. Let $X$, $Y$, and $Z$ be the input, output, and internal variables of $C_{SS}$, respectively. The sequence $S_2$ contains states on variables $X \cup Z \cup \{a\}$. The state of $S_2$ at an odd time point $t$ evaluates $X$ like the state at time $(t-1)/2$ of $S_1$ and all other variables $Z \cup \{a\}$ to false. This is enough for proving that $S_1 \preceq (S_2/2)$.

The states at an even time point have $a$ true, the value of $X$ as the next state, and the value of $Z$ that results from evaluating $C_{SS}$ with the values of $X$ as input. There exists a next state representation of polynomial size of this sequence because the successor of each state can be computed in polynomial time.

The LTL formula $L$ having only this sequence $S_2$ as a model is as follows. The initial state $s_0$ is translated into an LTL formula enforcing the state at time 0 to have the same values of $s_0$ on the variables $X$ and true on $a$. The formula $\mathbf{G}(a \not\equiv \mathbf{X}a)$ makes the value of $a$ change at each state. Therefore $a$ is true at all even time points and false at all odd time points. The formulae $\mathbf{G}(\neg a \rightarrow \neg z_i)$ makes all $z_i$ false at all odd time points.

When $a$ is true, that is, on even time points, the formula $\mathbf{G}(a \rightarrow C_{SS}[y_i/\mathbf{XX}x_i])$ makes the variables $Z$ in the current state and the variables $X$ after two time steps being the result of evaluating the circuit on the current values of $X$. Finally, $\mathbf{G}(a \rightarrow (x_i \equiv \mathbf{X}x_i))$ makes the values of $x_i$ at even time points to remain the same in the next time point. $\square$

As a result of this theorem, the time point problem remains $\|\rightsquigarrow$NP-hard for sequences in the next state representation even if we restrict to sequences that are the only models of some LTL formulae.

We can show a similar result for the time/state representation.

**Theorem 23** *Every sequence of states $S_1$ in the next state representation can be translated in polynomial time into an LTL formula $L$ and another sequence $S_2$ in the next state representation that is the only model of $L$ and $S_1 \preceq (S_2/2)$.*

*Proof.* Let $\langle C_{TS}, N \rangle$ be the time/state representation of the sequence $S_1$. We use the variables $T$ to represent time points. Let $T$, $X$, and $Z$ be the input, output, and internal variables of $C_{SS}$, respectively.

The sequence $S_2$ contains states on variables $X \cup Z \cup T \cup \{a\}$. This sequence is as follows: the states at odd time points $t$ are determined by the condition $S_1 \preceq (S_2/2)$: the values of $X$ are that of the $(t-1)/2$-th element of the sequence $S_1$ and all other variables are false.

In a state at an even time point $t$, the variable $a$ is true, $X$ is equal the value in the next time point, $T$ represent the number $t/2$, and $Z$ are the values of the internal variables of $C_{TS}$ when the inputs are $T$. A time/state representation of this sequence can be produced from $C_{TS}$ because the state at each time point can be generated in polynomial time.

We show the LTL formula $L$ whose only model is this sequence. A first subformula enforces the values of $T$ in the initial state to be all false and $a$ to be true. The formula $\mathbf{G}(a \not\equiv \mathbf{X}a)$ makes $a$ change value at each step. The formulae $\mathbf{G}(a \to (x_i \equiv \mathbf{X}x_i))$ makes the value of $X$ to remain the same if $a$ is true; therefore, $X$ maintain its value from even to odd time points.

By replacing each $\mathbf{X}$ with $\mathbf{XX}$ and $\mathbf{G}(\ldots)$ with $\mathbf{G}(a \to (\ldots))$, the formula $COUNT_T$ enforces the values of $T$ to increase by one from one even time point to the next one. Therefore, the values of $T$ at an even time $t$ are the binary representation of $t/2$. Finally, the formula $\mathbf{G}(a \to C_{TS})$ enforces the right relationship between $T$, $Z$, and $X$ at even time points. □

These two theorems prove that the relationship between the next state and the time/state representations remains the same if we add the assumption that sequences are the only models of an LTL formula. Note, however, that there are LTL formulae that do not have models representable in the next state representation, e.g., $\neg a \wedge \mathbf{X}\neg a \wedge \mathbf{XX}a \wedge \mathbf{XXX}a$.

### 4.4.2 Bounding Matters

In this section we consider the second point of the program of analysis of PSPACE-complete problems: showing that there are formulae that have models but no model of polynomial size. In particular, we show that:

1. there are formulae having only models with an exponentially long period;

2. there are satisfiable formulae whose models cannot be represented in space $\leq k$;

3. the Polynomial Solution Property does not hold for LTL.

Showing that there exists LTL formulae having only exponentially long models is trivial: the formula $COUNT_X$ has only models composed of exactly $2^{|X|}$ states. The same formula can be used to prove that there are formulae that are satisfiable but do not have models that can be represented in space bounded by a number $k$. The bits of the models of $COUNT_X$ all change at some point. As a result, no model, in whichever succinct form, can be shorter than $k = n$.

These questions were already obvious from the beginning. A more significant question is whether any satisfiable formula has a model that can be represented in polynomial space. This is however in contrast with the theorems of the previous section and the complexity of the time point and the next state problems.

### 4.4.3 Unary Representation

We now consider the problem of checking whether an LTL formula has a succinct model of size at most $k$, where $k$ is an integer in unary representation. We first show the membership to PSPACE of the problem for both representations.

**Theorem 24** *Checking whether an LTL formula has a model of size bounded by $k$ in unary in either the next state or the time/state representation is in* PSPACE.

*Proof.* Each circuit $C_{SS}$ with $X$, $Y$, and $Z$ as input, output, and internal variables, respectively, can be translated into the circuit $\mathbf{G}C_{SS}[y_i/\mathbf{X}x_i]$ that enforces the values of $X$ in the next state to be computed using $C$ from the values of the current state. As a result, if a model is in the next state representation, then satisfaction of an LTL formula w.r.t. this model can be checked by conjoining the formula with $\mathbf{G}C_{SS}[y_i/\mathbf{X}x_i]$.

In the same way, we can translate a time/state representation of a model into an LTL formula. The formula $COUNT_T$ allows the variables $T$ to always represent the current time point. The formula $\mathbf{G}C_{TS}$ then makes the values of $X$ to be exactly the values as specified by the sequence, if $T$, $X$, and $Z$ are the input, output, and internal values of the circuit $C_{TS}$, respectively. □

The following theorem tells that the satisfiability of an LTL formula in a simple lexicographic model is PSPACE-hard. Since this model can be easily represented in both the next state and the time/state representation, this theorem proves the hardness of the problem for both representations.

**Lemma 3** *For any QBF there exists an LTL formula that is satisfied by the model that is lexicographic on its variables and evaluates all variables to false in the initial state if and only if the QBF is valid.*

*Proof.* We consider an arbitrary QBF in which the variables have all odd indexes and are in reversed order $Q_n x_n, \ldots, Q_3 x_3, Q_1 x_1 \,.\, F$. In order to check the validity of this QBF, we need to check the value of $F$ for each evaluation of the variables, but then we have to combine the results according to the quantifiers.

Let $X = \{x_n, \ldots, x_1\}$ be the set of variables of the QBF interleaved with variables $x_i$ of even index, which do not appear in the QBF. The values of these variables in the lexicographic model can be visualized as follows.

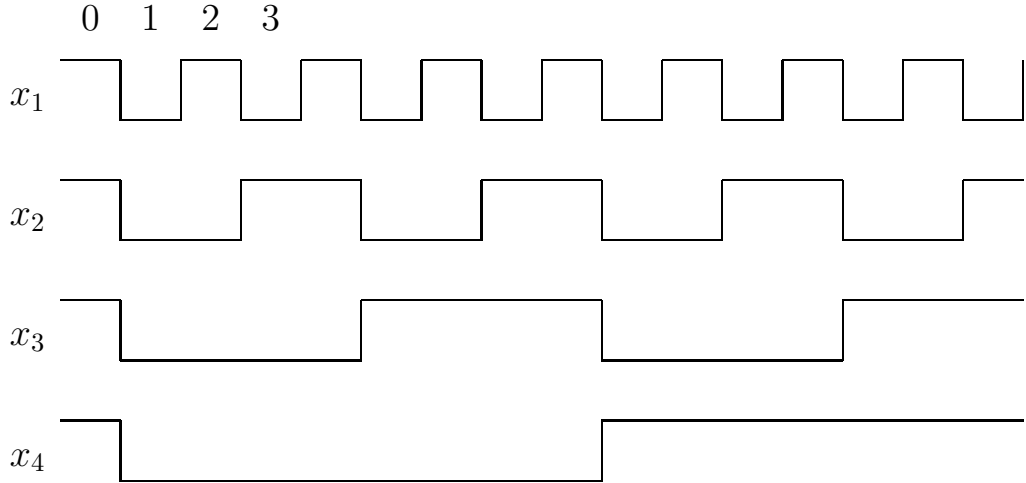

The LTL formula is built from the QBF recursively as follows: for each sub-formula $Q_i x_i \ldots Q_1 x_1 . F$ (in which $i$ is odd by assumption) we build an LTL formula $L_i$ that satisfies the following condition:

> for every truth evaluation $\omega_{n,i+2}$ of the variables $x_n, \ldots, x_{i+2}$
>
> > $Q_i x_i \ldots Q_1 x_1 . F \wedge \omega_{n,i+2}$ is valid
> >
> > if and only if

$L_i$ is true in the first point where $x_n, \ldots, x_{i+2} = \omega_{n,i+2}$.

By assumption, the QBF only contains the variables $x_i$ of odd index. The variables $x_i$ of even index are introduced because they are necessary for the proof.

We first define $L_1$ and then show that, given an arbitrary formula $L_i$ that satisfies the above condition, we can define the formula $L_{i+2}$ that satisfies the condition as well. If these formulae are all polynomial, then the formula $L_n$ is polynomial, and satisfied by the lexicographic model if and only if the QBF is valid.

Let us assume that $x_1$ is universally quantified. In order for the condition to be satisfied, $L_1$ must be true in the first state where $x_n, \ldots, x_3$ have values $\omega_{n,3}$ if and only if $F \wedge \omega_{n,3}$ is true for both $x_1 = $ false and $x_1 = $ true. This condition is equivalent to $F$ being valid in the first time point in which $x_n, \ldots, x_3 = \omega_{n,3}$ for the first time and in the following one. This condition can be expressed by the following LTL formula.

$$L_1 = F\mathbf{U}x_2$$

Consider for example the evaluation $x_n, \ldots, x_3 = $ false, $\ldots,$ false and the following figure. The variables $x_n, \ldots, x_3$ have values false, $\ldots,$ false for the first time in 0. The formula $L_1$ is valid in 0 if and only if $F$ is valid from 0 to the first time $x_2$ becomes true, excluded. Since $x_2$ becomes true for the first time in 2, $F$ must be true in 0 and 1. The assignment of the variables $x_n, \ldots, x_3$ in these time points are still false, $\ldots,$ false, while $x_1$ is false in 0 and true in 1. As a result, $L_1$ is valid in the first time point where $x_n, \ldots, x_3 = $ false, $\ldots,$ false if and only if $\forall x_1. F \wedge (x_n = $ false$) \wedge \cdots \wedge (x_3 = $ false$)$ is valid. The same holds for every time point in which $x_n, \ldots, x_3$ get new values.

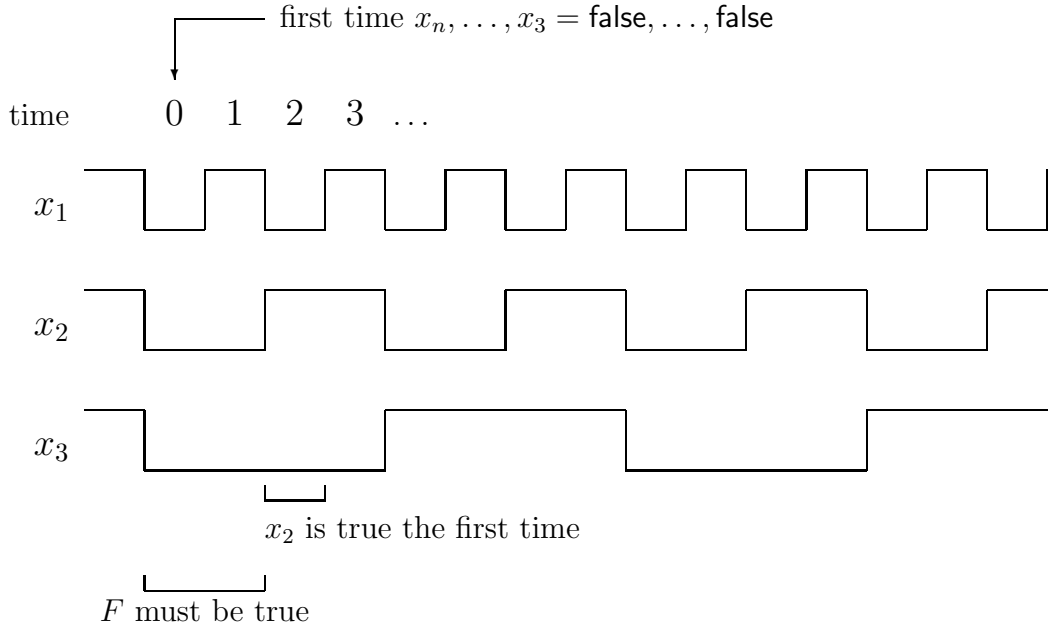

first time $x_n, \ldots, x_3 = \mathsf{false}, \ldots, \mathsf{false}$

$x_2$ is true the first time

$F$ must be true

If $x_1$ is existentially quantified, we can use the fact that $\exists x_1.F$ is equivalent to $\neg \forall x_1.\neg F$. Therefore, $L_1 = \neg((\neg F)\mathbf{U}x_2)$. In the example case of the first time point, this formula can be interpreted as "it is not true that $F$ is always false in time points 0 and 1."

Consider an arbitrary variable $x_i$ with odd index $i$. By assumption, $L_i$ is true in the first time point in which $x_n, \ldots, x_{i+2}$ assume a given value if and only if what remains of the QBF after setting that value is a valid formula. Assume that $x_{i+2}$ is universally quantified. A formula $L_{i+2}$ satisfying the condition above is the following one:

$$L_{i+2} = (\neg x_{i-1} \wedge \cdots \wedge \neg x_1 \to L_i)\mathbf{U}x_{i+1}$$

The following figure shows the value of the subformulae of $L_{i+2}$ on the diagram of evolution of the variables.

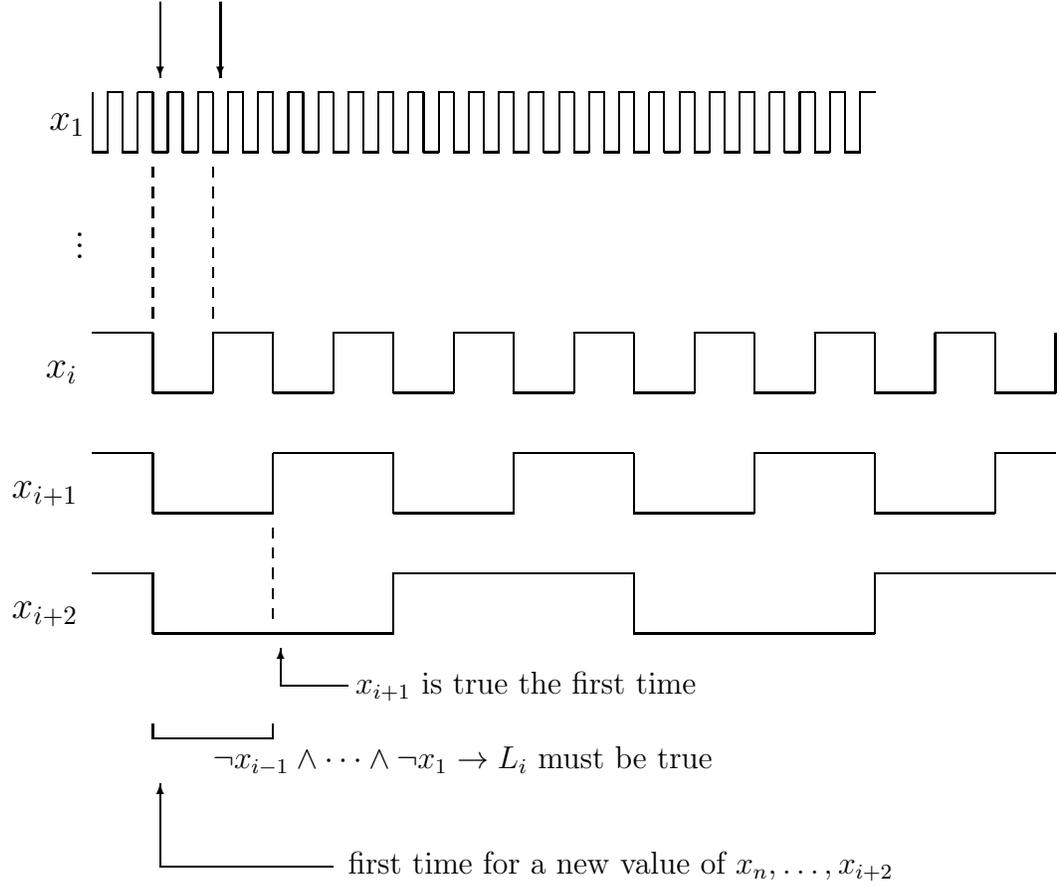

$\neg x_{i-1} \wedge \cdots \wedge \neg x_1$ is true in these points

$x_1$

$\vdots$

$x_i$

$x_{i+1}$

$x_{i+2}$

$x_{i+1}$ is true the first time

$\neg x_{i-1} \wedge \cdots \wedge \neg x_1 \to L_i$ must be true

first time for a new value of $x_n, \ldots, x_{i+2}$

Consider the first time point in which $x_n, \ldots, x_{i+2}$ assume a new value. Since $L_{i+2} = \text{formula}\mathbf{U}x_{i+1}$, the formula must be true in all time points in which $x_{i+1}$ remains false. In turns, the subformula $\neg x_{i-1} \wedge \cdots \wedge \neg x_1$ is true exactly in the first time point in which $x_i$ is false and the first time point in which it is true. By assumption $L_i$ tells the validity of the QBF that remains after $x_n, \ldots, x_{i+2}$ have been instantiated. As a result, $L_{i+2}$ tells the satisfiability of this QBF for both $x_i = \mathsf{false}$ and $x_i = \mathsf{true}$. This formula $L_{i+2}$ therefore satisfies the assumption. The case in which $x_i$ is quantified existentially is considered in a similar way.

By induction, $L_n$ is an LTL formula that is valid in the first time point if and only if the whole QBF is valid. $\square$

This theorem tells that, given a QBF, there exists an LTL formula such that the QBF is valid if and only if the LTL formula is valid in the model in which the variables are initially false and increase by one at each time step. This model is the only one satisfying the formula $COUNT_X \wedge \neg x_n \wedge \cdots \wedge \neg x_1$. The conjunction of these two formulae is satisfiable if and only if the QBF is valid.

**Theorem 25** *The problem of checking whether there exists a succinct model (either in the time/state or in the next state representation) of an LTL formula of size bounded by a number $k$ in unary notation is* PSPACE-*complete.*

*Proof.* Membership is proved in Theorem 24.

Hardness: by conjoining $COUNT_X \wedge \neg x_n \wedge \cdots \wedge \neg x_1$ with the formula that is satisfied by the lexicographic model iff the QBF is valid, we obtain a formula that is satisfiable if and only if the QBF is valid. The proof is completed by observing that the lexicographic model can be represented by a circuit (in either representation) of size $k = 4n$. □

### 4.4.4 Binary Representation

Every satisfiable formula has a model that can be represented in space exponential in the size of the formula. Since this exponential can be represented by a number $k$ in binary notation, the problem of checking the existence of a model of size bounded by a number $k$ in binary is at least as hard as the problem without the bound.

**Theorem 26** *The problem of establishing, given an LTL formula, of a model that can be represented in space bounded by an integer $k$ in binary notation is* PSPACE-*hard.*

The following lemma is needed for proving a result about the membership of this problem to some class.

**Lemma 4** *Given a satisfiable LTL formula, the initial and periodic parts of one of its models can be generated in polynomial space.*

*Proof.* The proof of membership to PSPACE by Sistla and Clarke [SE85] is based on guessing the subformulae of the given formula that are true in the current and in the next state. We start by guessing the sets of satisfiable subformulae $SF_1$ and $SF_2$ for the first and second state, respectively, and check the following conditions:

$$\begin{aligned}
A \wedge B \in SF_1 &\Leftrightarrow A \in SF_1 \text{ and } B \in SF_1 \\
\neg A \in SF_1 &\Leftrightarrow A \notin SF_1 \\
\mathbf{X}A \in SF_1 &\Leftrightarrow A \in SF_2 \\
A\mathbf{U}B \in SF_1 &\Leftrightarrow B \in SF_1 \text{ or} \\
&\quad A \in SF_1 \text{ and } A\mathbf{U}B \in SF_2
\end{aligned}$$

After this check has been done, we can forget about $SF_1$ and guess a new set of subformulae $SF_3$, etc. Some care has to be taken for the point where the period begins [SE85]. This nondeterministic PSPACE algorithm can be turned into deterministic thanks to Savitch theorem [Sav70].

We can produce all these sets $SF_i$ with a deterministic Turing machine as follows: we iterate over all possible pairs $SF_1$ and $SF_2$. This iteration only requires a polynomial amount of space. For each possible $SF_1$ and $SF_2$, we check whether there exists valid subsets $SF_3$, $SF_4$, etc. Since the existence of such sets $SF_3$, $SF_4$, etc. can be verified nondeterministically in polynomial space, it can also be done deterministically in polynomial space. Therefore, producing $SF_1$ only requires polynomial space. We can then deallocate the space needed by $SF_1$ and proceed in the same way for $SF_2$.

A specific state for each time point is generated because, if the formula $L$ contains a variable $x_i$, then $x_i$ is one of the subformulae of $L$. As a result, the above algorithm produces, for each time point, the set of variables $x_i$ that are true in that time point. □

A theorem similar to that of QBF and planning can then be established about the membership of the problem of checking the existence of a solution of size bounded by a number in binary for LTL.

**Theorem 27** *If the problem of existence of succinct LTL models of size bounded by a number $k$ in binary notation is not in* P, *then* P $\neq$ PSPACE.

*Proof.* We assume that P=PSPACE and prove that the problem is in P. By the above lemma, producing a model of a satisfiable formula can be done in polynomial space. Since P=PSPACE, this model can be produced in polynomial time as well. Therefore, there exists a circuit of polynomial size that represent such a model. The problem of existence of a model of an LTL formula can therefore be expressed as the existence of a circuit of size polynomial in that of the formula that represents a model of the formula. Checking this model with the formula is again in PSPACE, because one can easily convert a circuit into another LTL formula and then check its conjunction with the original formula.

Adding the bound does not change the complexity of the problem because checking the size of a polynomial-size circuit is easy. Therefore, the problem with the bound is in PSPACE, and is therefore in P. □

As a consequence of this theorem, if the problem is EXPTIME-hard, then it is not in P, and therefore P ≠ PSPACE.